\pdfoutput=1
\documentclass[pmlr,twocolumn,10pt]{jmlr} 

\usepackage{booktabs}
\usepackage{siunitx}
\usepackage[switch]{lineno}
\usepackage[markup=underlined]{changes}
\usepackage{xcolor}
\definecolor{AddedColor}{rgb}{1, 0, 0}
\setaddedmarkup{\textcolor{AddedColor}{\uline{#1}}}

\theorembodyfont{\upshape}
\theoremheaderfont{\scshape}
\theorempostheader{:}
\theoremsep{\newline}

\jmlrvolume{}
\jmlryear{2025}
\jmlrsubmitted{}
\jmlrpublished{}
\jmlrworkshop{Conference on Health, Inference, and Learning (CHIL) 2025}

\usepackage{amsmath} 
\usepackage{amssymb}
\usepackage{mathtools}
\usepackage{booktabs}
\usepackage{wrapfig}
\usepackage{placeins}

\usepackage{etoolbox}
\usepackage{outlines} 
\usepackage{tikz}
\usetikzlibrary{patterns}
\usetikzlibrary{shapes.geometric, arrows, calc}
\usepackage{mathrsfs}
\usepackage{subcaption}

\tikzstyle{observed_component} = [circle, text centered, draw=black, fill=gray!20, minimum size=1cm]
\tikzstyle{latent_component} = [circle, text centered, draw=black, minimum size=1cm]
\tikzstyle{arrow} = [->,>=stealth]
\tikzstyle{disparity_arrow} = [->,>=stealth, draw=red]

\usepackage{xcolor}

\newcommand{\severity}{{Z_t}}

\newcommand{\initialseverity}{{Z_0}}
\newcommand{\features}{{X_t}}
\newcommand{\rate}{{R}}
\newcommand{\visitindicator}{D_t}
\newcommand{\demographic}{A}
\newcommand{\avginitseverity}[1]{{\mu_{Z_0}^{(#1)}}}
\newcommand{\stdinitseverity}[1]{{\sigma_{Z_0}^{(#1)}}}
\newcommand{\pinnedgroup}{{a_0}}
\newcommand{\othergroup}{{a}}
\newcommand{\avgrate}[1]{{\mu_{R}^{(#1)}}}
\newcommand{\stdrate}[1]{{\sigma_{R}^{(#1)}}}
\newcommand{\featurescale}{F}
\newcommand{\featureintercept}{b}
\newcommand{\varfeaturenoise}{{\Psi}}
\newcommand{\featurenoise}{{\epsilon_t}}
\newcommand{\poissonintercept}{{\beta^{}_0}}
\newcommand{\poissonseveritycoeff}{{\beta^{}_Z}}
\newcommand{\poissongroupcoeff}[1]{{\beta_A^{(#1)}}}

\newcommand{\eventininterval}{E_t}
\newcommand{\eventinspecificinterval}{E_{t_i}}

\newcommand{\avginitseverityplain}{{\mu^{}_{Z_0}}}
\newcommand{\stdinitseverityplain}{{\sigma^{}_{Z_0}}}
\newcommand{\avgrateplain}{{\mu^{}_{R}}}
\newcommand{\stdrateplain}{{\sigma^{}_{R}}}
\newcommand{\poissongroupcoeffplain}{{\beta^{}_{\demographic}}}

\newcommand{\varinitseverityplain}{{{\sigma^{\ \ \ 2}_{Z_0}}}}
\newcommand{\varrateplain}{{{\sigma^{\ \ 2}_{R}}}}
\newcommand{\varinitseverityplaintilde}{{{\tilde{\sigma}^{\ \ \ 2}_{Z_0}}}}
\newcommand{\varrateplaintilde}{{{\tilde{\sigma}^{\ \ 2}_{R}}}}

\newcommand{\featurescaletilde}{{\tilde{F}}}
\newcommand{\featureintercepttilde}{{\tilde{b}}}
\newcommand{\varfeaturenoisetilde}{{\tilde{\Psi}}}
\newcommand{\avginitseveritytilde}[1]{{\tilde{\mu}_{Z_0}^{(#1)}}}
\newcommand{\stdinitseveritytilde}[1]{{\tilde{\sigma}_{Z_0}^{(#1)}}}
\newcommand{\avgratetilde}[1]{{\tilde{\mu}_{R}^{(#1)}}}
\newcommand{\stdratetilde}[1]{{\tilde{\sigma}_{R}^{(#1)}}}
\newcommand{\avginitseverityplaintilde}{{\tilde{\mu}^{}_{Z_0}}}
\newcommand{\stdinitseverityplaintilde}{{\tilde{\sigma}^{}_{Z_0}}}
\newcommand{\avgrateplaintilde}{{\tilde{\mu}^{}_{R}}}
\newcommand{\stdrateplaintilde}{{\tilde{\sigma}^{}_{R}}}
\newcommand{\poissonintercepttilde}{{\tilde{\beta}_0}}
\newcommand{\poissonseveritycoefftilde}{{\tilde{\beta}_Z}}
\newcommand{\poissongroupcoefftilde}[1]{{\tilde{\beta}_A^{(#1)}}}
\newcommand{\poissongroupcoeffplaintilde}{{\tilde{\beta}^{}_{\demographic}}}

\title[Learning Disease Progression Models That Capture Health Disparities]{Learning Disease Progression Models\\That Capture Health Disparities}

\author{
 \Name{Erica Chiang} \Email{ericachiang@cs.cornell.edu}\\
 \Name{Divya Shanmugam} \Email{divyas@cornell.edu}\\
 \addr Cornell Tech, Cornell University, USA
\AND
 \Name{Ashley N. Beecy} \Email{asb9028@nyp.org}\\
 \addr NewYork-Presbyterian, Weill Cornell Medical College, USA
 \AND
 \Name{Gabriel Sayer} \Email{gts2102@cumc.columbia.edu}\\
 \addr NewYork-Presbyterian, Columbia University Irving Medical Center, USA
\AND
 \Name{Deborah Estrin} \Email{destrin@cornell.edu}\\
 \Name{Nikhil Garg} \Email{ngarg@cornell.edu}\\
 \addr Cornell Tech, Cornell University, USA
 \AND
 \Name{Emma Pierson} \Email{emmapierson@berkeley.edu}\\
 \addr University of California, Berkeley, USA
}

\begin{document}

\maketitle

\begin{abstract}
Disease progression models are widely used to inform the diagnosis and treatment of many progressive diseases. However, a significant limitation of existing models is that they do not account for health disparities that can bias the observed data. To address this, we develop an interpretable Bayesian disease progression model that captures three key health disparities: certain patient populations may (1) start receiving care only when their disease is more severe, (2) experience faster disease progression even while receiving care, or (3) receive follow-up care less frequently conditional on disease severity. We show theoretically and empirically that failing to account for any of these disparities can result in biased estimates of severity (e.g., underestimating severity for disadvantaged groups). On a dataset of heart failure patients, we show that our model can identify groups that face each type of health disparity, and that accounting for these disparities while inferring disease severity meaningfully shifts which patients are considered high-risk. 
\end{abstract}

\paragraph*{Data and Code Availability}
This paper uses data from the NewYork-Presbyterian (NYP)/Weill Cornell Medical Center's 
electronic health record (EHR) system, which is not publicly available. Code for our model and all synthetic experiments can be found at \url{https://github.com/erica-chiang/progression-disparities}.

\section{Introduction}
\label{sec:intro}
In many settings, observed data is used to model the progression of a latent variable over time. Models of human aging use a person's physical and biological characteristics to model progression of their latent ``biological age'' \citep{pierson_inferring_2019}; models of infrastructure deterioration use inspection results to model progression of a system's latent overall health \citep{madanat_estimation_1995}; and disease progression models, which we focus on in this paper, use observed symptoms to model progression of a patient's latent severity of a chronic disease \citep{wang_unsupervised_2014}. Disease progression models can help predict a patient’s disease trajectory and thus personalize care, detect diseases at earlier stages, and guide drug development and clinical trial design \citep{mould_using_2007, romero_future_2015}. They have been applied to a wide variety of progressive diseases such as Parkinson's disease \citep{post_disease_2005}, Alzheimer's disease \citep{holford_methodologic_1992} and cancer \citep{gupta_extracting_2008}.

For the benefits of these models to apply to all patients equitably, it is crucial that they accurately describe progression for all patient populations. However, disease progression models have typically failed to account for the fact that systemic disparities in the healthcare process can bias the observed data that they are trained on. For example, disparities have been shown to arise along axes such as socioeconomic status \citep{weaver_forgoing_2010, miller_health_2017}, race \citep{yearby_racial_2018}, and proximity to care \citep{chan_geographic_2006, reilly_health_2021}. Accounting for such disparities is important because it can meaningfully shift estimates of disease progression. 
For intuition, imagine learning that a patient in the emergency room traveled three hours to get there; if their symptoms are ambiguous, this contextual information may increase our estimate of how severe their underlying condition is. 
Disease progression models have historically been unable to capture this type of context and, as we show, this can lead to biased estimates of severity. To address this, we propose a method for learning disease progression models that interpretably capture three well-documented health disparities:
\begin{enumerate}
    \item \textbf{Disparities in initial severity.} Certain patient groups may start receiving care only when their disease is more severe \citep{hu_screening_2024}. 
    \item \textbf{Disparities in disease progression rate.} Certain patient groups may experience faster disease progression, even while receiving care \citep{diamantidis_disparities_2021}.
    \item \textbf{Disparities in visit frequency.} Certain patient groups may visit healthcare providers for follow-up care less frequently, even at the same disease severity \citep{nouri_visit_2023}.
\end{enumerate}
A core technical challenge we address is designing a model that is flexible enough to capture all three disparities but still identifiable. Identifiability is necessary for accurate estimates of disparities and disease progression. As such, our key contributions are: (1) we develop an interpretable Bayesian model of disease progression that accounts for multiple types of disparities but remains provably identifiable from the observed data; (2) we prove and show empirically that failing to account for any of these three disparities leads to biased estimates of severity; and (3) we characterize fine-grained disparities in a heart failure dataset. Our model reveals that non-white patients have more severe heart failure and face multiple types of health disparities: Black and Asian patients tend to start receiving care at more severe stages of heart failure than do White patients, and Black patients see healthcare providers for heart failure 10\% less frequently than do White patients at the same disease severity level. Accounting for these disparities meaningfully shifts our estimates of disease severity, increasing the fraction of Black and Hispanic patients identified as high-risk. 

While we ground this work in healthcare, our method for learning progression models that account for disparities applies naturally to many other progression model settings where disparities are of interest, including infrastructure deterioration \citep{madanat_estimation_1995} and human aging \citep{pierson_inferring_2019}.
\section{Related Work}
\label{sec:related_work}

\paragraph{Disease progression modeling.} Disease progression models have been developed for many chronic diseases, including Parkinson's disease \citep{post_disease_2005}, Alzheimer's disease \citep{holford_methodologic_1992}, diabetes \citep{perveen_hybrid_2020}, and cancer \citep{gupta_extracting_2008}. A key feature of the progression models we consider, common in the machine learning literature, is that a latent severity $\severity$ progresses over time and gives rise to a set of observed symptoms $\features$. Models in this family include variants of hidden Markov models (HMMs) \citep{wang_unsupervised_2014, liu_efficient_2015, alaa_learning_2017, sukkar_disease_2012, jackson_multistate_2003} and recurrent neural networks (RNNs) \citep{choi_doctor_2016, lipton_learning_2017, lim_disease-atlas_2018, choi_retain_2016, ma_dipole_2017, kwon_retainvis_2019, alaa_attentive_2019}. This existing literature has not focused on modeling disparities; we extend it by proposing a new approach to disease progression modeling that can interpretably characterize and account for multiple types of health disparities. 

\paragraph{Health disparities.} Disparities have been documented in many parts of the healthcare process. Factors such as distance from hospitals \citep{reilly_health_2021}, distrust of the healthcare system \citep{laveist_mistrust_2009}, or lack of insurance \citep{venkatesh_association_2019} can result in underutilization of health services; biases in the judgements of healthcare providers can lead minority groups to receive later screening \citep{lee_disparities_2021}, fewer referrals \citep{landon_assessment_2021}, or generally worse care \citep{schafer_health_2016}; and issues such as limited health literacy or trust can create disparities in follow-through for appointments or the effectiveness of at-home care \citep{davis_physiologic_1968, brandon_legacy_2005}. 

The existing literature has shown that disparities emerge along the three axes that we capture in this paper: (1) how severe a patient's disease becomes before they start to receive care \citep{chen_clustering_2021, iqbal_differences_2015, hu_screening_2024}; (2) how quickly their latent severity progresses even while receiving care \citep{diamantidis_disparities_2021, suarez_racial_2018}; and (3) how likely they are to visit a healthcare provider at a given severity level \citep{nouri_visit_2023}. Our goal is to show how accounting for disparities along all three of these axes improves the severity estimates of disease progression models, while also learning more fine-grained descriptions of disparities. 

\paragraph{Capturing disparities with machine learning.} We build upon a large body of past work that uses machine learning as a tool to capture and address health disparities, including models that estimate the relative prevalence of underreported medical conditions \citep{shanmugam_quantifying_2021}, improve risk prediction for patients with missing outcome data \citep{balachandar_domain_2023}, evaluate the impact of race corrections in risk prediction \citep{zink_race_2023}, 
assess disparate impacts of AI in healthcare \citep{chen_can_2019}, and quantify disparities in the performance of clinical prediction tasks \citep{zhang_hurtful_2020}. The closest work to our own is \cite{chen_clustering_2021}, which develops a clustering algorithm that accounts for the fact that some patients do not come in (and are therefore not observed) until later in their disease progression. While their work addresses one form of data bias that can arise due to health disparities, it differs from our own in two ways: it does not specifically document or study health disparities, and it focuses on clustering patients as opposed to modeling disease severity or progression. Our work proposes a model for capturing three types of health disparities in the disease progression setting in order to learn precise descriptions of multiple disparities and make severity estimates that exhibit less bias than existing disease progression models. 
\begin{figure}[ht]
  \begin{center}
  \vskip -0.1in
\centerline{\includegraphics[width=0.4\textwidth]{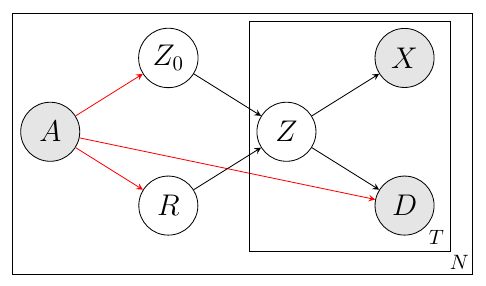}}
    \caption{\textbf{Disease progression generative model.} Plate diagram captures $N$ patients over $T$ timesteps. Shaded nodes indicate observed features: demographics $A^{(i)}$, visit indicator $\visitindicator^{(i)}$, and symptoms $\features^{(i)}$ (only observed when $\visitindicator^{(i)} = 1$). Unshaded nodes indicate latent variables: a patient's initial severity $\initialseverity^{(i)}$, rate of progression $\rate^{(i)}$, and severity $\severity^{(i)}$. Red arrows indicate dependencies capturing health disparities.}
    \label{fig:model_dag}
    \end{center}
    \vskip -0.3in
\end{figure}
\section{Model}
\label{sec:model}
We build on a standard setup for disease progression modeling, in which each patient has an underlying latent disease severity $\severity$ that progresses over time and gives rise to a set of observed features $\features$.

We characterize each patient's severity $\severity \in \mathbb{R}$ at time $t$ by their \textit{initial severity} $\initialseverity$ at their first observation (which we denote as $t=0$) and their \textit{rate of progression} $\rate$ after that point: 
\begin{equation*}
    \severity = \initialseverity + \rate \cdot t
\end{equation*} 
If a patient visits a healthcare provider at time $t$, we observe some recorded set of \textit{features} $\features \in \mathbb{R}^d$ (e.g., lab results, imaging, symptoms). At any given visit, a clinician does not necessarily observe or record all features---we model the features that \textit{are} observed as a noisy function of the patient's latent severity $\severity$: 
\begin{equation*}
    \features = f(\severity) + \epsilon_{t}
\end{equation*}
\begin{equation*}
    \epsilon_{t} \sim \mathcal{N}(0, \varfeaturenoise)
\end{equation*}

where the diagonal covariance matrix $\varfeaturenoise \in \mathbb{R}^{d \times d}$ parameterizes feature-specific noise (accounting for both measurement error and variation in how the patient's physical state can fluctuate day-to-day). In our experiments, we specifically instantiate $f$ as a linear function $f(\severity) = \featurescale \cdot \severity + \featureintercept$, where $\featurescale \in \mathbb{R}^d$ is a feature-specific scaling factor and $\featureintercept \in \mathbb{R}^d$ is a feature-specific intercept, but our approach extends to more general parametric forms for $f$. We constrain the first feature $\featurescale_0 > 0$ using domain knowledge; this restriction is necessary for identifiability because it restricts the mapping between features and severity \citep{shapiro1985identifiability}. We also observe a set of time values at which a patient visits a healthcare provider; we encode this with a binary indicator $\visitindicator\in \{0, 1\}$ that is equal to $1$ if a patient has a visit at time $t$ and $0$ otherwise.

\paragraph{Capturing disparities.} 

Our model captures the three types of health disparities discussed in \S \ref{sec:related_work} by allowing model parameters to vary as a function of a patient's demographic feature vector $\demographic$. For expositional clarity, we describe a setup where $\demographic$ encodes a single categorical label (e.g., a patient's race group), but our approach naturally extends to multiple categorical groupings or to continuous features. 

\begin{enumerate}
    \item \label{disparity1} \textbf{Disparities in initial severity.} Underserved patients may start receiving care only when their disease is more severe. We capture this by learning group-specific distributions of $\initialseverity$, a patient's disease severity at their first visit. For one group $\demographic=\pinnedgroup$, we pin $\initialseverity$ to be drawn from a unit normal distribution; this is a standard and necessary identifiability condition since it fixes the scale of $\severity$ \citep{shapiro1985identifiability}. For other groups $\othergroup$, we model 
    \begin{equation*}
         \initialseverity \sim \mathcal{N}\left(\avginitseverity{\othergroup}, \stdinitseverity{\othergroup}^2\right)
    \end{equation*}
    where $\avginitseverity{\othergroup}$ and $\stdinitseverity{\othergroup}$ are learned group-specific parameters. 
    
    \item \label{disparity2} \textbf{Disparities in disease progression rate.} Underserved patients may experience faster disease progression even while receiving care. We capture this by learning group-specific distributions of disease progression rate $\rate$:
    \begin{equation*}
        \rate \sim \mathcal{N}\left(\avgrate{\othergroup}, \stdrate{\othergroup}^2\right)
    \end{equation*}
    where $\avgrate{\othergroup}$ and $\stdrate{\othergroup}$ are learned group-specific parameters for each group $\othergroup$.
    
    \item \label{disparity3} \textbf{Disparities in visit frequency.} Underserved patients may visit healthcare providers for follow-up care less frequently at a given disease severity. We capture this by modeling patient visits as generated by an inhomogeneous Poisson process, parameterized by a time-varying rate parameter $\lambda_t$ that depends on both $\severity$ and $\demographic$:
    \begin{equation*}
        \log(\lambda_t) = \poissonintercept + \poissonseveritycoeff \cdot \severity + \poissongroupcoeff{\othergroup} 
    \end{equation*}
    where $\poissonseveritycoeff$ and $\poissonintercept$ are learned parameters for the entire population and $\poissongroupcoeff{\othergroup}$ is a learned group-specific parameter for each group $\othergroup$ (we pin $\poissongroupcoeff{\pinnedgroup} = 0$ for reference).
\end{enumerate}

\begin{table}[hbtp]
\centering
    {\begin{tabular}{@{}ll@{}}\toprule
    \textbf{Notation} & \textbf{Meaning} \\ 
    \midrule
    $\features$ & Observed features at time $t$\\
    $\visitindicator$ & Binary visit indicator for time $t$ \\
    $\demographic$ & Demographic features\\\midrule\midrule
    $\severity$ & Disease severity at time $t$ \\
    $\initialseverity$ & Initial severity \\
    $\rate$ & Disease progression rate\\
    $\featurescale$ & Severity-feature matrix \\
    $\featureintercept$ & Feature intercepts\\
    $\Psi$ & Feature covariance matrix\\
    $\avginitseverityplain,\stdinitseverityplain$ &  Group-specific mean and sd of $\initialseverity$ \\
    $\avgrateplain,\stdrateplain$ & Group-specific mean and sd of $\rate$\\
    $\lambda_t$ & Visit rate at time $t$\\
    $\poissonintercept$ & Visit rate intercept \\ 
    $\poissonseveritycoeff$ & Visit rate $\severity$ coefficient\\
    $\poissongroupcoeffplain$ & Visit rate $\demographic$ coefficient\\
    \bottomrule
    \end{tabular}} 
    \caption{\textbf{Summary of notation.} Observed data are listed above the double horizontal line.}
    \label{table:notation}
\end{table}

Overall, our model parameters (on which we place weakly informative priors) are the parameters shared across groups $\big\{\featurescale$, $\featureintercept$, $\varfeaturenoise$, $\poissonintercept$, $\poissonseveritycoeff\big\}$, and the group-specific parameters  $\big\{\avginitseverity{\othergroup}$, $\stdinitseverity{\othergroup}$, $\avgrate{\othergroup}$, $\stdrate{\othergroup}$, $\poissongroupcoeff{\othergroup}\big\}$. We learn posterior distributions over these parameters from our observed data $\big\{\features, \visitindicator, \demographic\big\}$ using Hamiltonian Monte Carlo, a standard algorithm for Bayesian inference \citep{betancourt_conceptual_2018}, as implemented in Stan \citep{carpenter_stan_2017}. Figure \ref{fig:model_dag} summarizes the data generating process and Table \ref{table:notation} summarizes the notation for our model.

\paragraph{Model discussion.} Our model makes several common assumptions. First, we model event frequency with a Poisson process; this is a common approach, including in work that seeks to capture disparities in event frequency~\citep{liu2024quantifying,kurashima2018modeling}. Second, we model progression as linear over time, a common approach for learning interpretable characterizations of trajectories \citep{holford_methodologic_1992, kimkoPredictionOutcomePhase2000, pierson_inferring_2019}. Finally, our assumption of a linear relationship between the latent state $\severity$ and feature values $\features$ is also standard; for example, factor analysis makes this assumption. While our linearity assumptions may limit our model's ability to capture some nuances of disease progression, they allow the model to interpretably capture progression trends over time; interpretability is especially valuable in our setting, allowing us to directly compare quantities like initial severity and progression rate across patient subgroups. 

\section{Theoretical analysis}
\label{sec:theory}
In this section, we prove two main theoretical results. First, we show that our model is \emph{identifiable}, a necessary condition for its parameters to be estimated from the observed data. Interpreting these parameter estimates is what allows us to quantify disparities. Second, we prove that failing to account for disparities produces \textit{biased estimates of severity}. We summarize proof strategies in the main text and provide formal proofs in Appendices \ref{app:proofs-of-identifiability}
 and \ref{app:proofs-of-bias}.

\subsection{Identifiability}
We show that our model is identifiable, meaning different sets of parameters yield different observed data distributions \citep{bellman_structural_1970}, which is necessary to correctly estimate model parameters from the observed data. Learning a model of progression that is \textit{flexible} enough to characterize multiple disparities but \textit{still identifiable} is a core challenge our work addresses. In fact, if we added one more dependence on $\demographic$---in particular, adding an arrow from $\demographic$ to $X$ in Figure \ref{fig:model_dag}---the model would no longer be identifiable (without a shared interpretation across groups of how features map to severity, it would be impossible to identify disparities in disease progression). 

\begin{theorem}\label{claim:identifiability}
All model parameters are identified by the observed data distribution $P(\features, \visitindicator \mid \demographic)$.
\end{theorem}

As mentioned in \S \ref{sec:model}, the distribution of initial severity $\initialseverity$ is pinned to a unit normal for one demographic group $\pinnedgroup$. This pinned distribution reduces the number of unknown latent parameters for group $\pinnedgroup$, allowing us to show that $\{\featurescale, \featureintercept, \varfeaturenoise\}$ are identified by $P(\features \mid \demographic = \pinnedgroup)$. Having identified these, we show that the parameters $\{\avginitseverity{a}, \stdinitseverity{a}, \avgrate{a}, \stdrate{a}\}$ are identified by $P(\features \mid \demographic=\othergroup)$ for all groups $\othergroup$. Finally, we show that given the previously identified parameters, $\{\poissonintercept, \poissonseveritycoeff\}$ are identified by $P(\visitindicator \mid \demographic=\pinnedgroup)$ and $\{\poissongroupcoeff{\othergroup}\}$ is identified by $P(\visitindicator \mid \demographic=\othergroup)$ for all other groups $\othergroup$. We provide a full proof in \S\ref{app-subsec:proof-identifiability}.

\subsection{Bias in models that do not account for disparities}
Next we show that, when any of the health disparities we discuss are present, a model that does not account for group-specific disparities will produce \textit{biased estimates} of severity---i.e., $\mathbb{E}[\severity \mid \features, \visitindicator] \neq \mathbb{E}[\severity \mid \features, \visitindicator, \demographic=\othergroup]$. 
These theoretical results hold whenever the model dependencies are encoded by the graph in Figure \ref{fig:model_dag}, a more general assumption than our full parametric model. For each proof, we analyze the effect of one disparity---e.g., for disparities in initial severity, we assume that $P(\initialseverity \mid \demographic)$ differs across groups---while keeping other distributions constant across groups. The results hold in the presence of multiple disparities as long as the existing disparities disfavor or favor the same group, so as to not cancel each other out in their effects.

To quantify disparities, we use the strict Monotone Likelihood Ratio Property (MLRP) to reason about the probability density functions of initial severity and progression rate for certain groups, relative to the overall population \citep{karlin_1956}:
\begin{definition}
    Two distributions characterized by probability density functions $f(x)$ and $g(x)$ have the strict monotone likelihood ratio property in $x$ if $\frac{f(x)}{g(x)}$ is a strictly increasing function of $x$. 
\end{definition}
Intuitively, this means that as some variable $x$ ($\initialseverity$ or $\rate$, in our case) gets larger, it is more likely to be drawn from $f$ than $g$. The MLRP is a widely-used assumption across many settings \citep{gaebler_simple_2024, anwar_alternative_2006, CHEMLA2019104929}; the normal, exponential, binomial, and Poisson families all have this property. For brevity, we say ``$f(x)$ strictly MLRPs $g(x)$'' to mean that $f(x)$ and $g(x)$ satisfy the strict MLRP in $x$. We now prove for each disparity that any model failing to account for the disparity will produce biased estimates of severity.

\begin{theorem}\label{claim:proof-initseverity-disparity}
A model that does not take into account disparities in initial disease severity $\initialseverity$ will underestimate the disease severity of groups with higher initial severity and overestimate that of groups with lower initial severity. Specifically, if $P(\initialseverity \mid \demographic=\othergroup)$ strictly MLRPs $P(\initialseverity)$ for some group $\othergroup$, then $\mathbb{E}[\severity \mid \features] < \mathbb{E}[\severity \mid \features, \demographic=\othergroup]$. Similarly, if $P(\initialseverity)$ strictly MLRPs $P(\initialseverity \mid \demographic=\othergroup)$ for some group $\othergroup$, then $\mathbb{E}[\severity \mid \features] > \mathbb{E}[\severity \mid \features, \demographic=\othergroup]$.
\end{theorem} 

We prove this by showing that $P(\initialseverity \mid \features, \demographic=\othergroup)$ strictly MLRPs $P(\initialseverity \mid \features)$, which implies that $\mathbb{E}[\severity \mid \features, \demographic=\othergroup] > \mathbb{E}[\severity \mid \features]$; \S\ref{app-subsec:proof-initseverity-disparity} provides a full proof. 

\begin{theorem}\label{claim:proof-severityrate-disparity}
Suppose disease severity progresses linearly at some rate $\rate$. A model that does not take into account disparities in $\rate$ will underestimate the disease severity of groups with higher progression rates and overestimate that of groups with lower progression rates. Specifically, if $P(\rate \mid \demographic=\othergroup)$ strictly MLRPs $P(\rate)$ for some group $\othergroup$, then $\mathbb{E}[\severity \mid \features] < \mathbb{E}[\severity \mid \features, \demographic=\othergroup]$. Similarly, if $P(\rate)$ strictly MLRPs $P(\rate \mid \demographic=\othergroup)$ for some group $\othergroup$, then $\mathbb{E}[\severity \mid \features] > \mathbb{E}[\severity \mid \features, \demographic=\othergroup]$.
\end{theorem}
We use a similar proof technique as for Theorem \ref{claim:proof-initseverity-disparity} and provide a full proof in \S\ref{app-subsec:proof-severityrate-disparity}. 

Finally, we analyze disparities in visit frequency. For this portion of the theoretical analysis, we consider discrete, non-infinitesimal time intervals (with the $i$-th interval starting at time $t_i$, where $t_0=0$) and whether or not a patient visited at any point during each interval. We introduce a new random variable $\eventinspecificinterval$ to indicate whether the patient has any visits during the time interval starting at $t_i$ (i.e., whether $\visitindicator=1$ for some $t$ in $[t_i, t_{i+1})$). We show that conditioned on $\eventininterval$, the group with lower visit frequency has higher expected severity at the beginning of the interval. 

\begin{theorem}\label{claim:proof-visit-disparity}
A model that does not take into account disparities in visit frequency conditional on disease severity will underestimate the disease severity of groups with lower visit frequency conditional on severity and overestimate the disease severity of groups with higher visit frequency conditional on severity. Specifically, assume that $P(\eventininterval=1\mid\severity)$ is strictly monotone increasing in $\severity$, $\lim\limits_{\severity\rightarrow -\infty} P(\eventininterval=1\mid\severity) = 0$, and $\lim\limits_{\severity\rightarrow \infty} P(\eventininterval=1\mid\severity) = 1$. Then if some group $\othergroup$ has a lower probability of visiting a healthcare provider at any given severity level---that is, $P(\eventininterval=1\mid\severity=z, \demographic=\othergroup) = P(\eventininterval=1\mid\severity=z  - \alpha(z))$ for all $z$, where $\alpha(z)$ is a positive function of $z$---then $\mathbb{E}[\severity \mid \eventininterval] < \mathbb{E}[\severity \mid \eventininterval,\demographic=\othergroup]$. Similarly, if $P(\eventininterval=1\mid\severity=z, \demographic=\othergroup) = P(\eventininterval=1\mid\severity=z + \alpha(z))$ for all $z$, where $\alpha(z)$ is a positive function of $z$, then $\mathbb{E}[\severity \mid \eventininterval] > \mathbb{E}[\severity \mid \eventininterval,\demographic=\othergroup]$.
\end{theorem}

We prove this by directly reasoning about the expected value of $\severity$ when conditioning on group versus not; we provide a full proof in \S\ref{app-subsec:proof-visit-disparity}. 

Overall, these results convey the importance of accounting for disparities in disease progression models: it is fundamentally not possible to make well-calibrated estimates of severity without accounting for group differences in initial severity, progression rate, and visit frequency. 
\section{Synthetic experiments}
\label{sec:simulation}

In this section, we validate our model and theoretical results in synthetic data simulations. We generate synthetic datasets according to the modeling assumptions in \S\ref{sec:model} (with parameter values for each dataset drawn randomly from each parameter's prior distribution). For each dataset, we generate simulated data for two separate groups, differing in their distributions of initial severity, progression rate, and visit frequency (characterized by different $\avginitseverityplain$, $\avgrateplain$, and $\poissongroupcoeffplain$, respectively).

\subsection{Identifiability and Severity Estimation} 
\label{subsec-identifiability}

\begin{figure}[ht]
  \begin{center}
    \centerline{\includegraphics[width=0.34\textwidth]{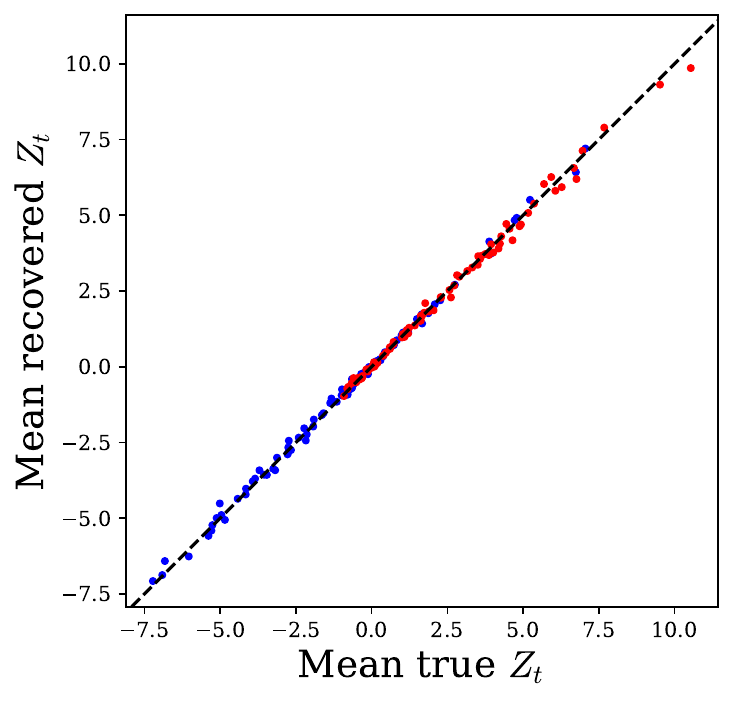}}
    \caption{\textbf{Well-calibrated severity estimates.} Each dot shows the mean true vs. mean recovered severity values for one group in a given simulation trial. Groups depicted in red tend to be underserved compared to groups depicted in blue. Our full model produces accurate and well-calibrated severity estimates (estimates lie near dotted $y=x$ line).}
    \label{fig:full_model_severity_recovery}
    \end{center}
    \vskip -0.3in
\end{figure}

We first verify Theorem \ref{claim:identifiability} in simulations, showing that when we fit our model on synthetic data, it accurately recovers the true data-generating parameters. We do this by examining the concordance between the model's estimated parameters and the true, latent parameter values, a common approach in past work \citep{chang_mobility_2021, pierson_inferring_2019}. We find high correlation between the true parameters and our model's posterior mean estimates (mean Pearson's $r$ 0.96 across all parameters; median 0.99), and good calibration (mean linear regression slope  1.0; median 1.0 when fit without an intercept term). We provide scatterplots of true versus estimated values for all parameters in Appendix \ref{app:simulations}. We also see that our model's mean \emph{severity estimates} for each group are highly correlated and well-calibrated with ground truth, despite underlying differences in group severity distributions and visit rates (Figure \ref{fig:full_model_severity_recovery}).

\subsection{Bias in models that do not account for disparities}
We now demonstrate in simulation that failing to account for disparities can lead to biased severity estimates, consistent with Theorems \ref{claim:proof-initseverity-disparity}, \ref{claim:proof-severityrate-disparity}, and \ref{claim:proof-visit-disparity}. In each trial, we use the same data to fit four models: our full model, which accounts for all disparities, plus three ablated models that each fail to account for one of the disparities (initial severity, progression rate, visit frequency). To characterize the resulting bias of failing to account for each type of disparity, we compute the average error in severity estimates (mean inferred estimate minus mean true severity) of each model, broken down by group. For each ablated model and trial, we define the ``underserved group'' to be the one that is underserved with respect to the specific disparity that the model fails to capture. When evaluating our full model, we define the ``underserved group'' to be the one with higher initial severity.

As seen in Table \ref{table:bias_simulation}, the models that do not account for disparities produce biased estimates: while our full model achieves average error across all trials of -0.02 for underserved patient groups and 0.01 for other patient groups, the ablated models all have negative error for underserved patients (underestimated severity) and positive error for other patients (overestimated severity). The ablated models also produce severity estimates that are less correlated with true severity.

\renewcommand{\arraystretch}{1.2}
\begin{table*}\centering
\small
\begin{tabular}{@{}lrrrrcrrr@{}}\toprule
&& \multicolumn{3}{c}{\textbf{Model that fails to account for disparities in...}}\\
\cmidrule{3-5}
& \textbf{Full model} & \emph{Initial severity} & \emph{Progression rate} & \emph{Visit frequency}\\
\midrule
Underserved group bias & -0.02 & -0.89 & -0.04 & -0.37 \\
Non-underserved group bias & 0.01 & +1.02 & +0.20 & +0.33 \\
Underserved group correlation & 1.00 & 0.72 & 0.99 & 0.94 \\
Non-underserved group correlation & 1.00 & 0.73 & 0.90 & 0.97 \\
\bottomrule
\end{tabular}
\caption{\textbf{Failing to account for disparities produces biased estimates of severity $\severity$.} We compare severity estimates from our full model to three ablated models that each fail to account for one of the three health disparities. While our full model produces accurate, well-calibrated severity estimates, each ablated model underestimates severity for the underserved group and overestimates it for the other group. The ablated model estimates are also \emph{less correlated} with the true severity values.}
\label{table:bias_simulation}
\end{table*}

\section{Modeling health disparities in heart failure progression}
\label{sec:case_study}
Finally, we fit our model on a real-world dataset of heart failure patients in the NewYork-Presbyterian hospital system. Heart failure is a progressive disease that affects many people, requires both specialty and preventive care \citep{colucci_2020_overview}, and has known health disparities \citep{lewsey_racial_2021}, making it a natural application setting for our model. In \S\ref{sec:heart_failure_data} we summarize the dataset, and in \S\ref{sec:model_validation} we confirm that our model can learn meaningful low-dimensional representations of disease severity by evaluating its reconstruction and predictive performance compared to standard baselines. In \S\ref{subsec:disparity-analysis} we present our main results: we interpret our model's learned parameters to provide precise descriptions of health disparities in our setting, and we show that (as our theory predicts) failing to account for these disparities meaningfully shifts severity estimates. 

\subsection{Data}
\label{sec:heart_failure_data}
Our data comes from the NewYork-Presbyterian (NYP)/Weill Cornell Medical Center's electronic health record (EHR) system from 2012-2020. We analyze a cohort of $N=2,942$ patients who (1) have a specific subtype of heart failure (heart failure with reduced ejection fraction), to ensure our cohort can be described by a single progression model, and (2) are likely to receive most of their cardiology care in the NYP system, to ensure we can reasonably estimate when they receive care. 

Observed feature data $\features$ for each patient includes four types of measurements: left ventricle ejection fraction (LVEF), brain natriuretic peptide (BNP), systolic blood pressure (SBP), and heart rate (HR). LVEF and BNP have strong clinical associations with heart failure severity (in terms of both underlying physiological health and observed symptoms) \citep{murphy_heart_2020}. SBP and HR are less informative (more prone to fluctuation and changes not related to heart failure), but they are still expected to show general trends over time as a patient's heart failure progresses. Since we must pin the sign of at least one scaling factor $F$ for identifiability, and decreasing LVEF is strongly associated with increasing severity in the heart failure subtype we study, we pin the sign of the scaling factor between severity $\severity$ and LVEF values ($F^{}_{\text{LVEF}} < 0$).

Measurements close in time are often from the same hospital visit, so we combine measurements within the same week (which has the additional benefits of increasing the speed of model fitting and allowing us to focus on capturing longer-term changes in disease severity). Specifically, for each week, we average together all measurements of the same type and treat any measurements as if they occurred at the beginning of the week.
We then capture disparities across four self-reported race/ethnicity groups: White non-Hispanic patients, Black non-Hispanic patients, Hispanic patients, and Asian non-Hispanic patients (which we will hereby describe as White, Black, Hispanic, and Asian subgroups). A full description of our data processing can be found in Appendix \ref{app:real-data}.

\subsection{Model validation}
\label{sec:model_validation}
We first confirm that our model accurately fits the data: we verify that the model's inferred parameters are consistent with medical knowledge (\S\ref{sec:consistency_with_medical_knowledge}) and compare the model's reconstruction and predictive performance to standard baselines (\S\ref{sec:predictive_performance}). We then show in \S\ref{subsec:disparity-analysis}, as our primary result, that our model provides insight into disparities in disease progression. 

\subsubsection{Consistency with medical knowledge}
\label{sec:consistency_with_medical_knowledge}

\begin{figure}
    \centering
        \includegraphics[height=4.5cm]{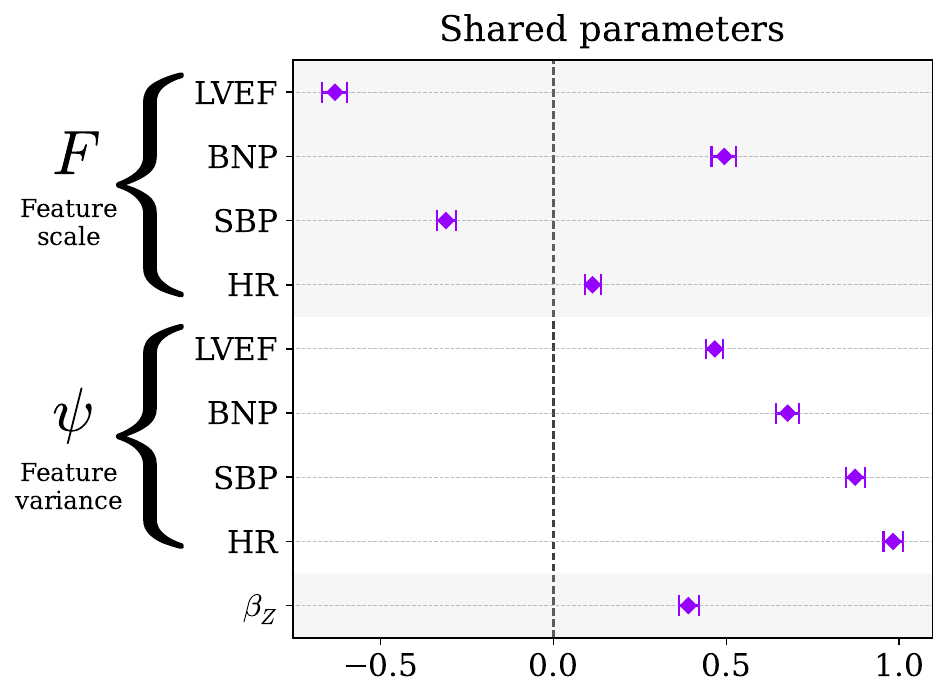}

        \includegraphics[height=4.5cm]{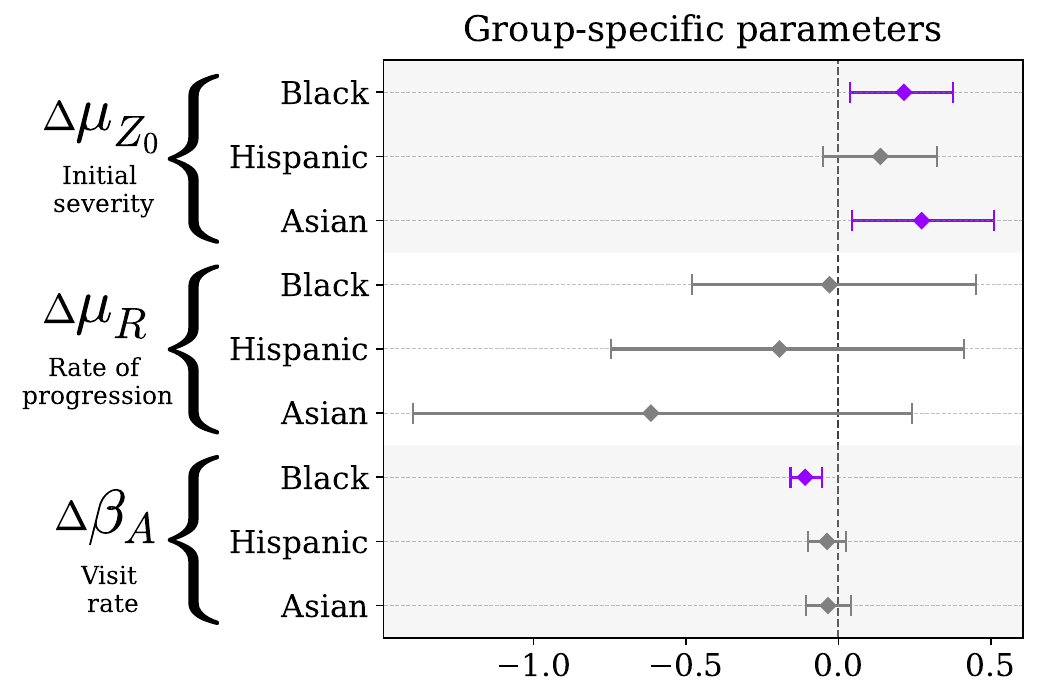}

    \caption{\textbf{Inferred model parameters with 95\% confidence intervals.} Shared parameters (top) are consistent with medical knowledge of heart failure progression. Group-specific parameters (bottom) are plotted as differences compared to White patients, so confidence intervals that are non-overlapping with 0 (colored in purple) indicate significant racial/ethnic differences in parameters.}
    \label{fig:coeff_plot}
    \vspace{-0.15in}
\end{figure}

Figure \ref{fig:coeff_plot} plots our model's inferred parameters, all of which are consistent with existing medical knowledge.\footnote{For succinctness, Figure \ref{fig:coeff_plot} plots only the model parameters of primary interest for interpreting our model (omitting, for example, estimated intercepts for each feature); a similar coefficient plot with all learned parameters is shown in Figure \ref{fig:all_coeff_plot}.} Specifically, (1) the model correctly learns that BNP and HR tend to increase with heart failure severity ($F^{}_{\text{BNP}}, F^{}_{\text{HR}}> 0$), while SBP tends to decrease ($F^{}_{\text{SBP}} < 0$) \citep{murphy_heart_2020}; (2) the model learns larger variance parameters for SBP and HR values ($\psi$), correctly inferring that these features are less informative about heart failure progression than BNP and LVEF \citep{murphy_heart_2020}; and (3) the model estimates that $\poissonseveritycoeff > 0$, meaning it learns that patients with higher disease severity tend to see healthcare providers more frequently, as expected.

\subsubsection{Reconstruction and predictive performance}
\label{sec:predictive_performance}

We next evaluate the model's ability to reconstruct and predict patient features $\features$. Because the model represents each patient visit in terms of a scalar severity $\severity$, we do not expect the model to perfectly reconstruct or predict the multi-dimensional $\features$; rather, we hope for predictions that correlate significantly with $\features$. Consistent with this, when fit on 3 years of data per patient, our model's predicted feature values correlate with true values both in- and out-of-sample. As we would hope, the model best represents the features that are known to be most informative about heart failure progression---LVEF ($r=0.81$ in-sample, $r=0.51$ out-of-sample) and BNP ($r=0.62$ in-sample, $r=0.31$ out-of-sample)---as opposed to the less-informative features SBP ($r=0.42$ in-sample, $r=0.24$ out-of-sample) and HR ($r=0.17$ in-sample, $r=0.03$ out-of-sample); all p-values besides HR out-of-sample $<0.001$.

To provide a more detailed assessment of performance, we evaluate our model's ability to \emph{reconstruct} patient features $\features$ in-sample and \emph{predict} $\features$ out-of-sample, in comparison to seven standard baselines. While predicting and reconstructing $\features$ is not the primary goal of our model, the model performs generally well relative to these baselines, validating its ability to meaningfully represent the data.

All of the baselines are designed to reconstruct or predict only the feature values $\features$; our model can also predict the occurrence of patient visits ($\visitindicator$), but in order to provide a direct comparison of reconstruction and predictive performance, we compare only the feature prediction aspect of our model (so we do not fit any models using $\visitindicator$ data) in this subsection. We use mean absolute percentage error (MAPE) to report a normalized measure of error across features. 

\paragraph{Reconstruction performance.} We compare our model's reconstruction performance to that of two standard \textit{dimensionality reduction baselines}: principal component analysis (PCA) and factor analysis (FA).
We compare our model to two variants of each. First, we compare our model to PCA and FA fit at the \emph{visit level}: one component per patient visit, analogous to our model's $\severity$. Second, we compare our model to PCA and FA fit at the \emph{patient level}: two components for each patient, to capture the trajectory of feature values as we do with $\initialseverity$ and $\rate$. We describe the implementation of these baselines with more detail in Appendix \ref{app:model-evaluation}. Because both PCA and FA require input vectors of consistent size, all models are fit on feature values from the first three visits per patient. Compared to all baselines, we achieve equivalent or better reconstruction performance across all features, and better performance on the more informative features (Table \ref{table:reconstruction}).

\paragraph{Predictive Performance.} 
We also compare our model's predictive performance to that of three standard \textit{timeseries forecasting baselines}: (1) a linear regression for each patient and feature; (2) a quadratic regression for each patient and feature; and (3) predicting values equal to those at the last timestep in training data. For this comparison, all models are fit on feature values from the first three years of data per patient, and we evaluate predictive performance on all remaining visits. While prediction is not the primary goal of our model (and models with relatively low predictive performance can still provide useful insights on disparities \citep{piersonAlgorithmicApproachReducing2021}), these results serve as an additional validation of our model's ability to meaningfully represent the data. Our model outperforms both linear regression and quadratic regression on all features. Our model has slightly higher MAPE than latest timestep, which is a widely-used, strong baseline for pure predictive performance \citep{hyndman2018forecasting}; latest timestep does not, however, provide any insight into disparities or even patterns of progression over time (Table \ref{table:prediction}). 

\subsection{Analysis of disparities}
\label{subsec:disparities}
We now discuss three main findings from fitting our model on the heart failure data. We learn that (1) Black patients tend to have higher disease severity than White patients; (2) our model learns precise descriptions of health disparities and finds that disparities of multiple types exist in our setting; and (3) failing to account for the existing disparities meaningfully shifts severity estimates for all racial/ethnic groups. This analysis is descriptive and does not require evaluating held-out performance, so models are fit on all available data. 

\begin{figure}
    \centering\includegraphics[height=3.65cm]{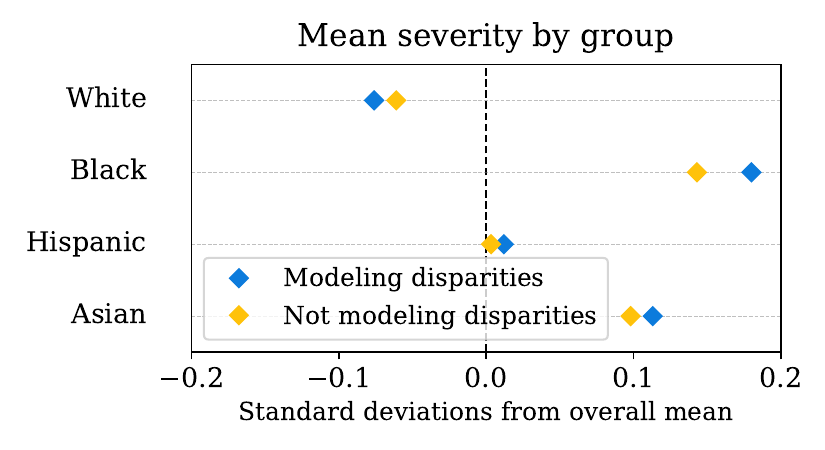}
    
    \includegraphics[height=3.65cm]{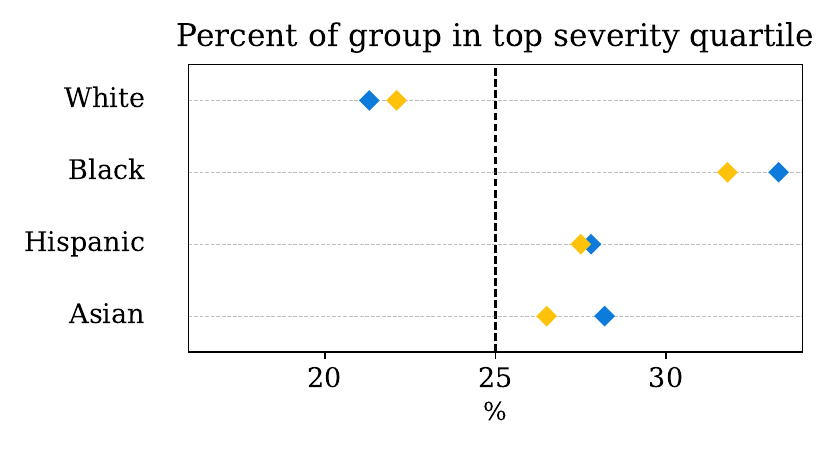}
    \caption{\textbf{Accounting for disparities leads to less biased severity estimates.} We visualize the improvement of our full model (blue) over one that does not account for disparities but is otherwise the same (yellow) in two ways. On the top, we show each group's average difference from the overall mean severity, normalized by the overall standard deviation of severity. On the bottom, we capture the portion of each group that is identified as ``high-risk'' (top quartile of disease severity).}
    \label{fig:estimate-shifts}
    \vspace{-0.2in}
\end{figure}

\paragraph{Black patients have higher disease severity.} 
As seen in Figure \ref{fig:estimate-shifts}, our model infers that Black patients have significantly higher disease severity than White patients ($p < 0.05$, computed by cluster bootstrapping at the patient-level; Hispanic and Asian patients also have higher inferred mean severity than White patients, but the differences are not statistically significant).

\label{subsec:disparity-analysis}
\paragraph{Model parameters capture fine-grained disparities.}
As seen in Figure \ref{fig:coeff_plot} (bottom), our model infers that Black and Asian patients have significantly higher initial severity than do White patients (inferred average initial severity $\avginitseverityplain$ for Black and Asian patient groups is greater than for White patients by 0.22 and 0.27, respectively). 
To contextualize the magnitude of these disparities, if all patients progressed at the average learned progression rate across the entire population, Black patients' first heart failure visit would occur 3.0 years later in their disease progression than White patients', and Asian patients' first visit would occur 3.8 years later. 
We also observe that $\poissongroupcoeffplain$ for Black patients is significantly lower than that of White patients, indicating that Black patients visit healthcare providers 10\% less frequently than White patients with the same disease severity. We describe these calculations in Appendix \ref{app:disparities-estimates}.

\paragraph{Accounting for disparities increases estimated severity for all non-white patient groups.} We compare severity estimates from our model to those of an ablated model that does not account for disparities (but is otherwise identical) and find that this meaningfully shifts severity estimates (Figure \ref{fig:estimate-shifts} top): while both models learn that non-white patients tend to have higher severity, the ablated model produces higher severity estimates for White patients and lower estimates for other groups ($p<0.001$ for all groups, computed by cluster bootstrapping at the patient-level). This is consistent with our theoretical results.

To highlight some implications of these shifted severity estimates, we look at each model's ranking of patient severity values and profile of ``high-risk'' visits: visits where inferred severity lies in the top quartile (25\%) of all visits. The ablated model is less likely to rank Black or Hispanic patient visits as high risk (Figure \ref{fig:estimate-shifts} bottom; $p<0.05$, computed by cluster bootstrapping at the patient-level), skewing the demographics of the high-risk patient cohort \textit{away} from groups that we know to have higher disease severity.

\section{Discussion}
\label{sec:discussion}
In this paper, we formalize three specific types of health disparities that bias observed health data: underserved patients may (1) first receive care only when their disease is more severe, (2) progress faster even while receiving care, or (3) receive care less frequently even at the same disease severity. We prove that failing to account for any of these disparities while learning disease progression can lead to biased estimates of severity, and we develop a disease progression model to capture all three disparities while provably retaining interpretability and identifiability. 

Our model can be used to make less biased severity estimates from patient health data \textit{and} to learn fine-grained descriptions of disparities in observational health data. Using a real-world heart failure dataset, we show that accounting for health disparities does indeed meaningfully shift severity estimates (by increasing the proportion of non-white patients identified as high-risk) and validate the model's ability to identify groups that face each type of health disparity. We thus urge future work in disease progression modeling to account for disparities in healthcare; we lay a foundation for doing so by developing a method to (1) make disease severity estimates that are accurate across diverse populations of patients and (2) learn interpretable estimates of distinct disparities that can inform future public health interventions. 

There are several natural directions for future work. First, beyond heart failure, our approach could be applied to many other progressive diseases, including Parkinson's \citep{post_disease_2005}, Alzheimer's \citep{holford_methodologic_1992}, diabetes \citep{perveen_hybrid_2020}, and cancer \citep{gupta_extracting_2008}, where it is possible that disparities manifest differently. Future work should similarly validate findings across multiple sites to assess generalizability of findings. To improve prediction and allow our model to more accurately capture rich medical data sources, another interesting technical direction is to extend our model to use additional data modalities (e.g., medical images) or more flexible progression models (e.g., non-linear trajectories), while retaining its interpretability and identifiability. Finally, our approach generalizes naturally to progression model settings beyond healthcare where disparities are of interest, including infrastructure deterioration \citep{madanat_estimation_1995} and human aging \citep{pierson_inferring_2019}; these would be interesting domains for future work. 

\paragraph*{Institutional Review Board (IRB)}
This study was approved by the institutional review boards at Weill Cornell Medicine (IRB \#22-06024904) and Columbia University Irving Medical Center (IRB \#AAAU1701). A waiver for informed consent was obtained.

\acks{EC is supported by NSF DGE \#2139899.
NG is supported by NSF CAREER \#2339427, and Cornell Tech Urban Tech Hub, Meta, and Amazon research awards.
EP is supported by a Google Research Scholar award, NSF CAREER \#2142419, a CIFAR Azrieli Global scholarship, a gift to the LinkedIn-Cornell Bowers CIS Strategic Partnership, the Abby Joseph Cohen Faculty Fund, an AI2050 Early Career Fellowship, and the Survival and Flourishing Fund.
This work was partially supported by funding from NewYork-Presbyterian for the NYP-Cornell Cardiovascular AI Collaboration, as well as funding from the MacArthur Foundation.
Thank you to
Gabriel Agostini,
Sidhika Balachandar,
Sarah Cen,
Serina Chang,
Evan Dong,
Albert Gong,
Sophie Greenwood,
Evelyn Horn,
Rishi Jha,
Chris Kelsey,
Konwoo Kim,
Mitchel Lang,
Benjamin Lee,
Zhi Liu,
Linda Lu,
Rajiv Movva,
Chidozie Onyeze,
Tom Reilly,
Jeffrey Ruhl,
Shuvom Sadhuka,
and
Naomi Tesfuzigta
for valuable conversations and feedback on this paper.}


\appendix
\newpage
\setcounter{figure}{0}
\renewcommand{\figurename}{Figure}
\renewcommand{\thefigure}{S\arabic{figure}}
\setcounter{table}{0}
\renewcommand{\tablename}{Table}
\renewcommand{\thetable}{S\arabic{table}}
\onecolumn

\section{Proof of Identifiability}
\label{app:proofs-of-identifiability}

\subsection{Proof of Theorem ~\ref{claim:identifiability}}
\label{app-subsec:proof-identifiability}
\paragraph{Theorem \ref{claim:identifiability}} \textit{All model parameters are identified by the observed data distribution $P(\features, \visitindicator \mid \demographic)$.}\\

\begin{proof} We want to show that each unique set of parameter assignments leads to a different distribution over the observed data. To do this, we divide our argument into four lemmas: 
\begin{lemma}
\label{app-subsec:proof-identifiability-a0-severity}
    Parameters $\featurescale, \featureintercept, \varfeaturenoise$ are identified by $P(\features \mid \demographic=\pinnedgroup)$.
\end{lemma}
\begin{quote}
\begin{proof}

We want to show that if two parameter sets $\{\featurescale, \featureintercept, \varfeaturenoise\}$ and $\{\featurescaletilde, \featureintercepttilde, \varfeaturenoisetilde\}$ yield the same observed data distribution $P(X_0 \mid A=\pinnedgroup)$, the parameter sets must be identical.

We first note that at $t=0$, we have $\severity = \initialseverity \sim \mathcal{N}(0, 1)$ for group $\pinnedgroup$. Then the mapping between severity and features 
$$X_0 = \featurescale \cdot Z_0 + \featureintercept + \featurenoise$$
$$\featurenoise \sim \mathcal{N}(0, \varfeaturenoise)$$

captures a factor analysis model with factor loading matrix $\featurescale$ and diagonal covariance matrix $\varfeaturenoise$. At $t=0$, the feature distribution for group $\pinnedgroup$ has  the standard factor analysis distribution~\citep{shapiro1985identifiability}: 
\begin{equation*}
    X_0 \sim \mathcal{N}(\featureintercept, \featurescale\featurescale^T + \varfeaturenoise).
\end{equation*}

Assuming the two sets of parameters map to distributions of $X_0$ with the same mean, it must hold that $\featureintercept = \featureintercepttilde$. Thus, parameter $\featureintercept$ is identified by data distribution $P(X_0 \mid A=\pinnedgroup)$.

Further, the covariance matrix of $X_0$ induced by each set of parameters must be the same: $\featurescale (\featurescale)^T  + \varfeaturenoise = \featurescaletilde (\featurescaletilde)^T + \varfeaturenoisetilde$. Element-wise equality of the covariance matrix gives us the following, where subscripts $i$ refer to the $i$-th element of each parameter vector:
\begin{equation}
\label{eq:off_diagonal_equality_constraint}
    \featurescale_i \featurescale_j = \featurescaletilde_i \featurescaletilde_j ~~\forall i, j, i\neq j
\end{equation}
\begin{equation}
\label{eq:diagonal_equality_constraint}
    (\featurescale_i)^2 + \varfeaturenoise_i  = (\featurescaletilde_i)^2 + \varfeaturenoisetilde_i 
\end{equation}

Using the equality constraint (\ref{eq:off_diagonal_equality_constraint}) for multiple pairs of indices, we have that for all assignments of distinct indices $i, j, k$:
\begin{equation}
\label{eq:combining_two_pairs}
    (\featurescale_i \featurescale_j = \featurescaletilde_i \featurescaletilde_j) \wedge (\featurescale_j \featurescale_k = \featurescaletilde_j \featurescaletilde_k) \implies \frac{\featurescaletilde_i}{\featurescale_i} = \frac{\featurescaletilde_k}{\featurescale_k}
\end{equation}
\begin{equation}
\label{eq:simplifying_one_pair}
    \featurescale_i \featurescale_k = \featurescaletilde_i \featurescaletilde_k \implies \frac{\featurescale_i}{\featurescaletilde_i} = \frac{\featurescaletilde_k}{\featurescale_k}
\end{equation}

Together, equations \ref{eq:combining_two_pairs} and \ref{eq:simplifying_one_pair} give us:

\begin{equation*}
    \frac{\featurescaletilde_i}{\featurescale_i} = \frac{\featurescale_i}{\featurescaletilde_i} \implies (\featurescaletilde_i)^2 = (\featurescale_i)^2 \implies \featurescale_i = \alpha \featurescaletilde_i
\end{equation*}
 
where $\alpha \in \{-1, +1\}$. Since we have fixed $\featurescale_0 > 0$ for \textit{all} factor loading matrices $\featurescale$, the sign of $\alpha$ is fixed:
\begin{equation}
\label{eq:equal_F}
    \featurescale_0 = \alpha \featurescaletilde_0 \implies \alpha = 1 \implies \featurescale_i = \featurescaletilde_i ~~\forall i \in [0, d),
\end{equation}

meaning we have identified $\featurescale$.

Lastly, using equations (\ref{eq:diagonal_equality_constraint}) and (\ref{eq:equal_F}) we get $\featurescale_i = \featurescaletilde_i \implies \varfeaturenoise_i = \varfeaturenoisetilde_i$. We have now shown that if two parameter sets induce the same distribution of $X$ at time $t=0$, they must have the same exact value assignments. Therefore $\featurescale, \featureintercept, \varfeaturenoise$ are identified by $P(\features \mid A=\pinnedgroup)$.
\end{proof}
\end{quote}

\begin{lemma}
\label{app-subsec:proof-identifiability-a-severity}
    Global parameters $\featurescale, \featureintercept, \varfeaturenoise$ and parameters $\avginitseverity{\othergroup}, \stdinitseverity{\othergroup}, \avgrate{\othergroup}, \stdrate{\othergroup}$ for each group $\othergroup$ are identified by $P(\features \mid \demographic)$.
\end{lemma}
\begin{quote}
\begin{proof}
By Lemma \ref{app-subsec:proof-identifiability-a0-severity}, we know that $\featurescale, \featureintercept, \varfeaturenoise$ are identified by $P(X_0 \mid A=\pinnedgroup)$. We want to show that for any group $\othergroup$, if two parameter sets $\{\avginitseverity{\othergroup}, \stdinitseverity{\othergroup}, \avgrate{\othergroup}, \stdrate{\othergroup}\}$ and $\{\avginitseveritytilde{\othergroup}, \stdinitseveritytilde{\othergroup}, \avgratetilde{\othergroup}, \stdratetilde{\othergroup}\}$ yield the same observed data distribution $P(X_t \mid A=\othergroup)$, the parameter sets must be identical. In this proof we consider an arbitrary group $\othergroup$ and omit the $(\othergroup)$ superscript for brevity.

We model the following:
\begin{equation*}
    \initialseverity \sim \mathcal{N}\left(\avginitseverityplain, \varinitseverityplain\right)
\end{equation*}
\begin{equation*}
    \rate \sim \mathcal{N}\left(\avgrateplain, \varrateplain\right)
\end{equation*}
\begin{equation*}
    \severity = \initialseverity + \rate \cdot t \implies \severity \sim \mathcal{N}\left(\avgrateplain \cdot t + \avginitseverityplain, \varrateplain \cdot t^2 + \varinitseverityplain\right)
\end{equation*}
\begin{equation}
\label{eq:factor_analysis_model_2}
    \features = \featurescale \cdot \severity + \featureintercept + \featurenoise \text{, where } \featurenoise \sim \mathcal{N}(0, \varfeaturenoise)
\end{equation}

We see that equation (\ref{eq:factor_analysis_model_2}) captures a factor analysis model with factor loading matrix $\featurescale$ and diagonal covariance matrix $\varfeaturenoise$, meaning
\begin{equation*}
    \features \sim \mathcal{N}(\featureintercept + \featurescale(\avgrateplain \cdot t + \avginitseverityplain), \featurescale(\varrateplain \cdot t^2 + \varinitseverityplain)\featurescale^T + \varfeaturenoise).
\end{equation*}

Recalling that $\featurescale_0 > 0$, we first consider $t=0$, where $X_0 \sim \mathcal{N}(\featureintercept + \featurescale\avginitseverityplain, \featurescale(\varinitseverityplain)\featurescale^T + \varfeaturenoise)$. In order for the two parameter sets to map to distributions of $X_0$ with the same mean, it must be the case that  
\begin{equation*}
    \featureintercept + \featurescale \avginitseverityplain = \featureintercept + \featurescale \avginitseverityplaintilde \implies \avginitseverityplain = \avginitseverityplaintilde.
\end{equation*}

Further, for the two parameter sets to map to distributions with the same covariance matrix, it must hold that 
\begin{equation*}
    \featurescale(\varinitseverityplain)\featurescale^T + \varfeaturenoise = \featurescale(\varinitseverityplaintilde)\featurescale^T + \varfeaturenoise \implies \stdinitseverityplain = \stdinitseverityplaintilde
\end{equation*}

since we know $\stdinitseverityplain,  \stdinitseverityplaintilde > 0$. So we have identified $\avginitseverityplain$ and $\stdinitseverityplain$. We next consider any time $t\neq 0$. For the two parameter sets to map to distributions of $\features$ with the same mean, given that we have already shown $\avginitseverityplain$ must equal $\avginitseverityplaintilde$, it must hold that  
\begin{equation*}
    \featureintercept + \featurescale(\avgrateplain \cdot t + \avginitseverityplain) = \featureintercept + \featurescale(\avgrateplaintilde \cdot t + \avginitseverityplaintilde) \implies \avgrateplain = \avgrateplaintilde.
\end{equation*}

For the two parameter sets to map to distributions with the same covariance matrix, given that we have already shown $\stdinitseverityplain$ must equal $\stdinitseverityplaintilde$, it must hold that  \begin{equation*}
    \featurescale(\varrateplain \cdot t^2 + \varinitseverityplain)\featurescale^T + \varfeaturenoise = \featurescale(\varrateplaintilde \cdot t^2 + \varinitseverityplaintilde)\featurescale^T + \varfeaturenoise \implies \stdrateplain = \stdrateplaintilde
\end{equation*}

since $\stdrateplain, \stdrateplaintilde > 0$. Thus we have shown that for any group $\othergroup$, group-specific values of $\avginitseverityplain, \stdinitseverityplain, \avgrateplain, \stdrateplain$ are identified by $P(\features \mid \demographic=\othergroup)$.

\end{proof}
\end{quote}

\begin{lemma}
\label{app:proof-poisson-universal-identifiability}
    Global parameters $\poissonintercept, \poissonseveritycoeff$ and the parameter $\poissongroupcoeff{\othergroup}$ for each group $\othergroup$ are identified by $P(\visitindicator \mid A)$.
\end{lemma}
\begin{quote}
\begin{proof}
We want to show that if two parameter sets  \{$\poissonintercept$, $\poissonseveritycoeff, \poissongroupcoeff{\othergroup}$\} and \{$\poissonintercepttilde$, $\poissonseveritycoefftilde, \poissongroupcoefftilde{\othergroup}$\} yield the same observed data distribution $P(\visitindicator \mid A=\othergroup)$, the parameter sets must be identical. Unless otherwise specified, we consider an arbitrary group $\othergroup$ and omit the $(\othergroup)$ superscript for brevity. We also assume $\avgrateplain \neq 0$, since in general the severity of a progressive disease should change over time and it does not make sense to learn progression in the case that it does not. 

Each event when a patient visits the hospital ($\visitindicator = 1$) is generated by an inhomogeneous Poisson process parameterized by $\lambda_t$, where $\log(\lambda_t) = \poissonintercept + \poissonseveritycoeff \cdot \severity  + \poissongroupcoeffplain$. 

In order for two data distributions to have identical $P(\visitindicator \mid A=\othergroup)$ they must have identical expected rates $\mathbb{E}_{\initialseverity, \rate}[\lambda_t]$: $\mathbb{E}_{\initialseverity, \rate}[\lambda_t]$ is the expected rate of events (across the population) at time $t$---if two distributions have a different expected rate of events at any time $t$, then $P(\visitindicator \mid \demographic=\pinnedgroup)$ must differ at that point in time as well. Thus if two sets of parameters \{$\poissonintercept$, $\poissonseveritycoeff$, $\poissongroupcoeffplain$\} and \{$\poissonintercepttilde$, $\poissonseveritycoefftilde$, $\poissongroupcoeffplaintilde$\} yield the same observed data distribution $P(\visitindicator \mid \demographic=\othergroup)$, they must also generate the same observed values $\mathbb{E}_{\initialseverity, \rate}[\lambda_t]$ at all timesteps $t$. We finish the proof by showing that this holds only if \{$\poissonintercept$, $\poissonseveritycoeff$, $\poissongroupcoeffplain$\} $=$ \{$\poissonintercepttilde$, $\poissonseveritycoefftilde$, $\poissongroupcoeffplaintilde$\}. 
\begin{align*}
    \mathbb{E}_{\initialseverity, \rate}[\lambda_t] &= \int\int\lambda_t \cdot P(\initialseverity) \cdot P(\rate)\ d\initialseverity d\rate
\end{align*}

By Lemma \ref{app-subsec:proof-identifiability-a-severity}, we know that $\avginitseverityplain, \stdinitseverityplain, \avgrateplain, \stdrateplain$ are identified by $P(\features \mid \demographic)$. Then
\begin{gather*}
    P(\initialseverity) = \frac{1}{\sqrt{2\pi(\stdinitseverityplain)^2}} \exp\left(-\frac{(\initialseverity-\avginitseverityplain)^2}{2(\stdinitseverityplain)^2}\right)\\
    P(\rate) = \frac{1}{\sqrt{2\pi(\stdrateplain)^2}} \exp\left(-\frac{(\rate-\avgrateplain)^2}{2(\stdrateplain)^2}\right)\\
\end{gather*}
\begin{equation}
\label{eq:lambda_expectation}
    \mathbb{E}_{\initialseverity, \rate}[\lambda_t] = \exp(f(\poissonintercept, \poissonseveritycoeff, \poissongroupcoeffplain, t))
\end{equation}
\begin{equation*}
    \text{where } f(\poissonintercept, \poissonseveritycoeff, \poissongroupcoeffplain, t) = \left(\frac{(\poissonseveritycoeff\stdrateplain)^2}{2}\right)t^2 + (\poissonseveritycoeff\avgrateplain)t + \left(\poissonintercept + \frac{(\poissonseveritycoeff\stdinitseverityplain)^2}{2} + \poissonseveritycoeff\avginitseverityplain + \poissongroupcoeffplain\right)
\end{equation*}

The expression in \ref{eq:lambda_expectation} must be equal for \{$\poissonintercept$, $\poissonseveritycoeff$, $\poissongroupcoeffplain$\} and \{$\poissonintercepttilde$, $\poissonseveritycoefftilde$, $\poissongroupcoeffplaintilde$\} at all timesteps $t$. Since $\exp$ is an injective function, this means that $f(\poissonintercept, \poissonseveritycoeff, \poissongroupcoeffplain, t) = f(\poissonintercepttilde, \poissonseveritycoefftilde, \poissongroupcoeffplaintilde, t)$ for all $t$. By equality of polynomials, each of the individual polynomial coefficients must be equal must be equal for this to hold. 

We first consider the case for group $\pinnedgroup$, since we pin $\poissongroupcoeff{\pinnedgroup}$ at 0 as a reference for all other groups. Given that we have already identified $\avginitseverity{\pinnedgroup}, \stdinitseverity{\pinnedgroup}, \avgrate{\pinnedgroup}, \stdrate{\pinnedgroup}$, 
\begin{gather*}
    \left(\poissonintercept + \frac{(\poissonseveritycoeff\stdinitseverityplain)^2}{2} + \poissonseveritycoeff\avginitseverityplain\right) = \left(\poissonintercepttilde + \frac{(\poissonseveritycoefftilde\stdinitseverityplain)^2}{2} + \poissonseveritycoefftilde\avginitseverityplain\right) \implies \poissonintercept = \poissonintercepttilde
\end{gather*}

Now we return to our analysis of any arbitrary group $\othergroup$. Given that we have already identified $\avginitseverityplain, \stdinitseverityplain, \avgrateplain\neq 0, \stdrateplain$,
\begin{gather*}
    \poissonseveritycoeff\avgrateplain = \poissonseveritycoefftilde \avgrateplain \implies \poissonseveritycoeff = \poissonseveritycoefftilde \\
    \left(\poissonintercept + \frac{(\poissonseveritycoeff\stdinitseverityplain)^2}{2} + \poissonseveritycoeff\avginitseverityplain + \poissongroupcoeffplain\right) = \left(\poissonintercepttilde + \frac{(\poissonseveritycoefftilde\stdinitseverityplain)^2}{2} + \poissonseveritycoefftilde\avginitseverityplain + \poissongroupcoeffplaintilde\right) \implies \poissongroupcoeffplain = \poissongroupcoeffplaintilde
\end{gather*}

Thus we have shown that $\poissonintercept, \poissonseveritycoeff$, and $\poissongroupcoeff{\othergroup}$ for any group $\othergroup$ are identified by $P(\visitindicator \mid \severity, \demographic)$.

\end{proof}
\end{quote}
    
By showing that each parameter of the model is uniquely recovered from the observed data, we have proved that our model is identifiable.
    
\end{proof}

\section{Proofs of Bias}
\label{app:proofs-of-bias}
In this section, in order to capture the effect of failing to account for one disparity at a time, we consider the setting where everything between two groups is the same except for disparity of focus. It is clear to see from our analysis that these results hold even more generally---as long as all existing disparities disfavor or favor the same group (e.g. a disadvantaged group with respect to one disparity is not advantaged with respect to another, in which case the effects could cancel each other out), our proofs of bias will hold. Throughout our proofs, we assume that all PDFs and conditional PDFs have positive support over their entire domain, and that all PDFs are differentiable, a very reasonable assumption over our setting.  

\subsection{Theorem ~\ref{claim:proof-initseverity-disparity}}
\label{app-subsec:proof-initseverity-disparity}
\paragraph{Theorem \ref{claim:proof-initseverity-disparity}} \textit{A model that does not take into account disparities in initial disease severity $\initialseverity$ will underestimate the disease severity of groups with higher initial severity and overestimate that of groups with lower initial severity. Specifically, if $P(\initialseverity \mid \demographic=\othergroup)$ strictly MLRPs $P(\initialseverity)$ for some group $\othergroup$, then $\mathbb{E}[\severity \mid \features] < \mathbb{E}[\severity \mid \features, \demographic=\othergroup]$. Similarly, if $P(\initialseverity)$ strictly MLRPs $P(\initialseverity \mid \demographic=\othergroup)$ for some group $\othergroup$, then $\mathbb{E}[\severity \mid \features] > \mathbb{E}[\severity \mid \features, \demographic=\othergroup]$.}\\

\begin{proof}
We want to show that $\mathbb{E}[\severity \mid \features, \demographic=\othergroup] > \mathbb{E}[\severity \mid \features]$ when $P(\initialseverity\mid \demographic=\othergroup)$ strictly MLRPs $P(\initialseverity)$. We first show that $P(\initialseverity \mid \features=x, \demographic=\othergroup)$ strictly MLRPs $P(\initialseverity \mid \features)$ with respect to $\initialseverity$:
\begin{align*}
    \frac{\partial}{\partial \initialseverity}\left(\frac{P(\initialseverity \mid \features, \demographic=\othergroup)}{P(\initialseverity \mid \features)}\right) 
    &= \frac{\partial}{\partial \initialseverity}\left(\frac{\frac{P(\features \mid \initialseverity, \demographic=\othergroup)P(\initialseverity \mid \demographic=\othergroup)}{P(\features \mid \demographic=\othergroup)}}{\frac{P(\features \mid \initialseverity)P(\initialseverity)}{P(\features)}}\right) \tag{Bayes Rule} \\
    &= \frac{\partial}{\partial \initialseverity}\left(\frac{\frac{P(\initialseverity \mid \demographic=\othergroup)}{P(\features \mid \demographic=\othergroup)}}{\frac{P(\initialseverity)}{P(\features)}}\right) \tag{$\features \perp A \mid \initialseverity, R$} \\
    &= \frac{P(\features)}{P(\features \mid \demographic=\othergroup)} \cdot \frac{\partial}{\partial \initialseverity}\left(\frac{P(\initialseverity \mid \demographic=\othergroup)}{P(\initialseverity)}\right) \\
    &> 0 \tag{Disparity assumption}
\end{align*}

Since MLRP implies first-order stochastic dominance (FOSD) \citep{klemens_when_2007}, this proves that $P(\initialseverity \mid \features, \demographic=\othergroup)$ strictly FOSDs $P(\initialseverity \mid \features)$ and thus that $\mathbb{E}[\initialseverity \mid \features,\demographic=\othergroup] > \mathbb{E}[\initialseverity \mid \features]$. By linearity of expectation,
\begin{align*}
    &\mathbb{E}[\initialseverity \mid \features,\demographic=\othergroup] + \mathbb{E}[f(\rate, t) \mid \features, \demographic=\othergroup] > \mathbb{E}[\initialseverity \mid \features]  + \mathbb{E}[f(\rate, t) \mid \features], \quad\forall t \geq 0\\
    &\implies \mathbb{E}[\severity \mid \features,\demographic=\othergroup] > \mathbb{E}[\severity \mid \features]
\end{align*}
It is clear to see that this argument extends naturally to show that if a group tends to come in at \textit{earlier} disease stages than the rest of the population, that their severity will be overestimated: If there exists a group $\tilde{\othergroup}$ such that $P(\initialseverity)$ strictly MLRPs $P(\initialseverity \mid \demographic=\tilde{\othergroup})$ with respect to $\initialseverity$ and $\mathbb{E}[R \mid \features] \geq \mathbb{E}[R \mid \features, \demographic=\tilde{\othergroup}]$, then we will see that $\mathbb{E}[\severity \mid \features,\demographic=\tilde{\othergroup}] < \mathbb{E}[\severity \mid \features]$. Hence any model that does not take into account demographic disparities in initial disease severity levels at a patient's first visit will lead to biased estimates of severity.
\end{proof}

\subsection{Proof of Theorem ~\ref{claim:proof-severityrate-disparity}}
\label{app-subsec:proof-severityrate-disparity}

\paragraph{Theorem \ref{claim:proof-severityrate-disparity}} \textit{Suppose disease severity progresses linearly at some rate $\rate$. A model that does not take into account disparities in $\rate$ will underestimate the disease severity of groups with higher progression rates and overestimate that of groups with lower progression rates. Specifically, if $P(\rate \mid \demographic=\othergroup)$ strictly MLRPs $P(\rate)$ for some group $\othergroup$, then $\mathbb{E}[\severity \mid \features] < \mathbb{E}[\severity \mid \features, \demographic=\othergroup]$. Similarly, if $P(\rate)$ strictly MLRPs $P(\rate \mid \demographic=\othergroup)$ for some group $\othergroup$, then $\mathbb{E}[\severity \mid \features] > \mathbb{E}[\severity \mid \features, \demographic=\othergroup]$.}

$\rate$ is a patient's linear rate of progression, so we model a patient's severity over time as $\severity = f(\rate, t) + \initialseverity$, where $f$ is linearly increasing in $\rate$. \\ 

\begin{proof}
We want to show that $\mathbb{E}[\severity \mid \features, \demographic=\othergroup] > \mathbb{E}[\severity \mid \features]$ when $P(\rate\mid \demographic=\othergroup)$ strictly MLRPs $P(\rate)$. We first show that $P(\rate \mid \features, \demographic=\othergroup)$ strictly MLRPs $P(\rate \mid \features)$ with respect to $R$:
\begin{align*}
    \frac{\partial}{\partial R}\left(\frac{P(\rate \mid \features, \demographic=\othergroup)}{P(\rate \mid \features)}\right) 
    &= \frac{\partial}{\partial R}\left(\frac{\frac{P(\features \mid \rate, \demographic=\othergroup)P(\rate \mid \demographic=\othergroup)}{P(\features \mid \demographic=\othergroup)}}{\frac{P(\features \mid \rate)P(\severity=z_t)}{P(\features)}}\right) \tag{Bayes Rule} \\
    &= \frac{\partial}{\partial R}\left(\frac{\frac{P(\rate \mid \demographic=\othergroup)}{P(\features \mid \demographic=\othergroup)}}{\frac{P(\rate)}{P(\features)}}\right) \tag{$X \perp \demographic \mid \initialseverity, R$} \\
    &= \frac{P(\features)}{P(\features \mid \demographic=\othergroup)} \cdot \frac{\partial}{\partial R}\left(\frac{P(\rate \mid \demographic=\othergroup)}{P(\rate)}\right) \\
    &> 0 \tag{Disparity assumption}
\end{align*}

Since MLRP implies FOSD \citep{klemens_when_2007}, this also implies that $P(\rate \mid \features, \demographic=\othergroup)$ strictly FOSDs $P(\rate \mid \features)$. It follows directly that $\mathbb{E}[R \mid \features, \demographic=\othergroup] > \mathbb{E}[R \mid \features]$. By linearity of expectation, 
\begin{align*}
    &\mathbb{E}[f(R, t) + \initialseverity \mid \features, \demographic=\othergroup] > \mathbb{E}[f(R, t) + \initialseverity \mid \features], \quad\forall t > 0 \\
    &\implies \mathbb{E}[\severity \mid \features,\demographic=\othergroup] > \mathbb{E}[\severity \mid \features]
\end{align*}

It is clear to see that this argument extends naturally to show that if a group tends to progress \textit{more slowly} than the rest of the population, that their severity will be overestimated: if there exists a group $\tilde{\othergroup}$ such that $P(\rate)$ strictly MLRPs $P(\rate \mid \demographic=\tilde{\othergroup})$ with respect to $R$ and $\mathbb{E}[\initialseverity \mid \features] \geq \mathbb{E}[\initialseverity \mid \features, \demographic=\tilde{\othergroup}]$, then we will see that $\mathbb{E}[\severity \mid \features,\demographic=\tilde{\othergroup}] < \mathbb{E}[\severity \mid \features]$. Thus any model that does not take into account demographic disparities in patient progression rates will lead to biased estimates of severity.
\end{proof}

\subsection{Proof of Theorem ~\ref{claim:proof-visit-disparity}}
\label{app-subsec:proof-visit-disparity}

\paragraph{Theorem \ref{claim:proof-visit-disparity}} \textit{A model that does not take into account disparities in visit frequency conditional on disease severity will underestimate the disease severity of groups with lower visit frequency conditional on severity and overestimate the disease severity of groups with higher visit frequency conditional on severity. Specifically, assume that $P(\eventininterval=1\mid\severity)$ is strictly monotone increasing in $\severity$, $\lim\limits_{\severity\rightarrow -\infty} P(\eventininterval=1\mid\severity) = 0$, and $\lim\limits_{\severity\rightarrow \infty} P(\eventininterval=1\mid\severity) = 1$. Then if some group $\othergroup$ has a lower probability of visiting a healthcare provider at any given severity level---that is, $P(\eventininterval=1\mid\severity=z, \demographic=\othergroup) = P(\eventininterval=1\mid\severity=z  - \alpha(z))$ for all $z$, where $\alpha(z)$ is a positive function of $z$---then $\mathbb{E}[\severity \mid \eventininterval] < \mathbb{E}[\severity \mid \eventininterval,\demographic=\othergroup]$. Similarly, if $P(\eventininterval=1\mid\severity=z, \demographic=\othergroup) = P(\eventininterval=1\mid\severity=z + \alpha(z))$ for all $z$, where $\alpha(z)$ is a positive function of $z$, then $\mathbb{E}[\severity \mid \eventininterval] > \mathbb{E}[\severity \mid \eventininterval,\demographic=\othergroup]$.} \\

\begin{proof} Recall that for a given patient, $\eventininterval$ corresponds to the event where some visit occurs during the timestep starting at time $t$, and $P(\eventininterval=1\mid\severity=z)$ indicates the probability that the patient visits during the time period if their severity is $z$ at the beginning of this time period. We will first show that, when there is some positive $\alpha(z)$ for all $z$ such that $P(\eventininterval=1\mid\severity=z, \demographic=\othergroup) = P(\eventininterval=1\mid\severity=z - \alpha(z))$, it holds that $\mathbb{E}[\severity \mid \eventininterval=1,\demographic=\othergroup] > \mathbb{E}[\severity \mid \eventininterval=1]$. We will then show that this argument holds when conditioning on $\eventininterval=0$ as well---i.e., $\mathbb{E}[\severity \mid \eventininterval=0,\demographic=\othergroup] > \mathbb{E}[\severity \mid \eventininterval=0]$. \\

We first compute $\mathbb{E}[\severity \mid \eventininterval=1]$. Define $p(z) \coloneq P(\severity=z)$ and $F(z) \coloneq P(\eventininterval=1\mid\severity=z)$. By Bayes' rule we have: 
\begin{align*}
    P(\severity = z \mid \eventininterval=1) = \frac{p(z)F(z)}{\int_{-\infty}^{\infty} p(z)F(z)\ dz}
\end{align*}

By assumption, $F(z)$ is strictly monotone increasing in $z$, $\lim\limits_{z\rightarrow -\infty} F(z) = 0$, and $\lim\limits_{z\rightarrow \infty} F(z) = 1$. These are the properties of a CDF, so we can write $F(z)$ in terms of its corresponding PDF $f(z)$: i.e., $F(z) = \int_{-\infty}^{z}f(x)dx$. This yields:
\begin{align*}
    P(\severity = z \mid \eventininterval=1) 
    &= \frac{p(z)\int_{-\infty}^z f(x)\ dx}{\int_{-\infty}^{\infty} p(z) \int_{-\infty}^z f(x)\ dx\ dz} 
\end{align*}

Then we can write the expectation $ \mathbb{E}[\severity \mid \eventininterval=1]$ as:
\begin{align*}
    \mathbb{E}[\severity \mid \eventininterval=1] &= \int_{-\infty}^{\infty} P(\severity = z \mid \eventininterval=1) z\ dz \\
    &= \frac{\int_{-\infty}^{\infty}\int_{-\infty}^z p(z)f(x)z\ dx\ dz}{\int_{-\infty}^{\infty} \int_{-\infty}^z p(z) f(x)\ dx\ dz}
\end{align*}

Graphically, this corresponds to taking the expectation of $z$ when points are sampled from the blue region in Figure \ref{fig:bias-proof-fig1}, where the probability of sampling each point is proportional to $p(z)f(x)$.

\FloatBarrier
\begin{figure}[ht]
    \begin{center}
    \centerline{\includegraphics[width=0.35\columnwidth]{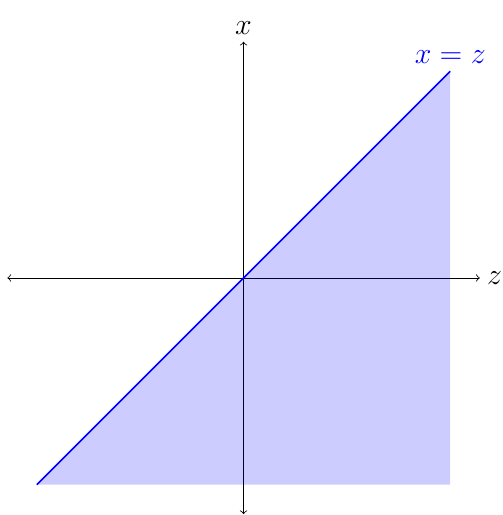}}
    \caption{We can calculate $\mathbb{E}[\severity \mid \eventininterval=1]$ by taking the expectation over the blue region, with each point having probability $p(z)f(x)$.}
    \label{fig:bias-proof-fig1}
    \end{center}
    \vskip -.2in
\end{figure}

\FloatBarrier

We next consider $\mathbb{E}[\severity \mid \eventininterval=1, \demographic=\othergroup]$, which yields an analogous expression. Define $p_a(z) := P(\severity=z \mid \demographic=\othergroup)$ and $F_a(z) = P(\eventininterval=1\mid\severity=z, \demographic=\othergroup)$. Since all groups have the same severity distribution, we see that $p_a(z) = p(z)$. Further, $F_a(z) = P(\eventininterval=1\mid\severity=z-\alpha(z)) = F(z-\alpha(z))$ by our disparity assumption. By Bayes' rule we have:
\begin{align*}
    P(\severity \mid \eventininterval=1, \demographic=\othergroup) &= \frac{p_a(z)F_a(z)}{\int_{-\infty}^{\infty} p_a(z) F_a(z)\ dz} \\
    &= \frac{p(z)F(z - \alpha(z))}{\int_{-\infty}^{\infty} p(z)F(z - \alpha(z))\ dz}
\end{align*}

As before, we write $F(z)$ in terms of its corresponding PDF $f(z)$, yielding:
\begin{align*}
    P(\severity \mid \eventininterval=1, \demographic=\othergroup) &= \frac{p(z)\int_{-\infty}^{z-\alpha(z)} f(x)\ dx}{\int_{-\infty}^{\infty} p(z)\int_{-\infty}^{z-\alpha(z)} f(x)\ dx\ dz}
\end{align*}

Finally, we can write the expectation $\mathbb{E}[\severity \mid \eventininterval=1, \demographic=\othergroup]$ as:
\begin{align*}
    \mathbb{E}[\severity \mid \eventininterval=1, \demographic=\othergroup] &= \int_{-\infty}^{\infty} P(\severity \mid \eventininterval=1, \demographic=\othergroup)z \ dz\\
    &=\frac{\int_{-\infty}^{\infty} \int_{-\infty}^{z-\alpha(z)} p(z)f(x)z \ dx\ dz}{\int_{-\infty}^{\infty} \int_{-\infty}^{z-\alpha(z)} p(z) f(x)\ dx\ dz}
\end{align*}

Now the region of $z$ and $x$ that we integrate over corresponds to the red region in Figure \ref{fig:bias-proof-fig2}; the crosshatched blue region corresponds to the region that we integrate over in our calculation of $\mathbb{E}[\severity\mid\eventininterval=1]$ but \textit{not} in our calculation of $\mathbb{E}[\severity\mid\eventininterval=1, \demographic=\othergroup]$ (such that the solid blue region in Figure \ref{fig:bias-proof-fig1} is the combination of the blue crosshatched and red regions in Figure \ref{fig:bias-proof-fig2}). Note that visualization of the red region assumes constant $\alpha(z)$, but the logical argument applies to any function $\alpha(z) > 0$.

\FloatBarrier
\begin{figure}[ht]
    \begin{center}
    \centerline{\includegraphics[width=0.35\columnwidth]{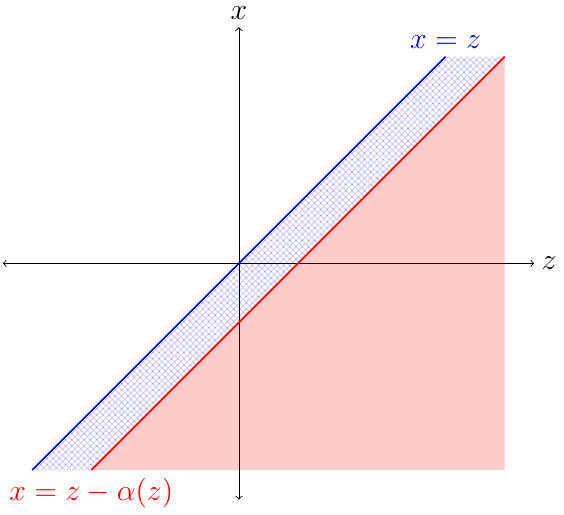}}
    \caption{We can calculate $\mathbb{E}[\severity \mid \eventininterval=1, \demographic=\othergroup]$ by taking the expectation over the red region, with each point having probability $p(z)f(x)$. The crosshatched blue region provides a comparison to the integration space for $\mathbb{E}[\severity \mid \eventininterval=1]$.}
    \label{fig:bias-proof-fig2}
    \end{center}
    \vskip -.2in
\end{figure}
\FloatBarrier

We see that at each value of $x$, the blue crosshatched region adds strictly positive weight to lower values of $z$; thus, the expectation of $z$ over the blue crosshatched region \textit{plus} red region must be lower than the expectation of $z$ over the red region itself. We therefore conclude that $\mathbb{E}[\severity\mid\eventininterval=1] < \mathbb{E}[\severity\mid\eventininterval=1, \demographic=\othergroup]$. 

The reasoning when conditioning on $\eventininterval=0$ is analogous. Since $P(\eventininterval=0\mid \severity=z) = 1 - F(z)$, we get the following expressions:
\begin{align*}
    \mathbb{E}[\severity \mid \eventininterval=0] &= \frac{\int_{-\infty}^{\infty}\int_{z}^{\infty} p(z)f(x)z\ dx\ dz}{\int_{-\infty}^{\infty} \int_{z}^{\infty} p(z) f(x)\ dx\ dz}\\
    \mathbb{E}[\severity \mid \eventininterval=0, \demographic=\othergroup] &=\frac{\int_{-\infty}^{\infty} \int_{z-\alpha(z)}^{\infty} p(z)f(x)z \ dx\ dz}{\int_{-\infty}^{\infty} \int_{z-\alpha(z)}^{\infty} p(z) f(x)\ dx\ dz}
\end{align*}

Graphically, $\mathbb{E}[\severity \mid \eventininterval=0]$ corresponds to taking the expectation of $z$ when points are sampled from the blue region in Figure \ref{fig:bias-proof-fig3}, where the probability of sampling each point is proportional to $p(z)f(x)$. $\mathbb{E}[\severity \mid \eventininterval=0, \demographic=\othergroup]$ corresponds to taking the expectation over the blue plus crosshatched red regions.

\FloatBarrier
\begin{figure}[ht]
    \begin{center}
    \centerline{\includegraphics[width=0.35\columnwidth]{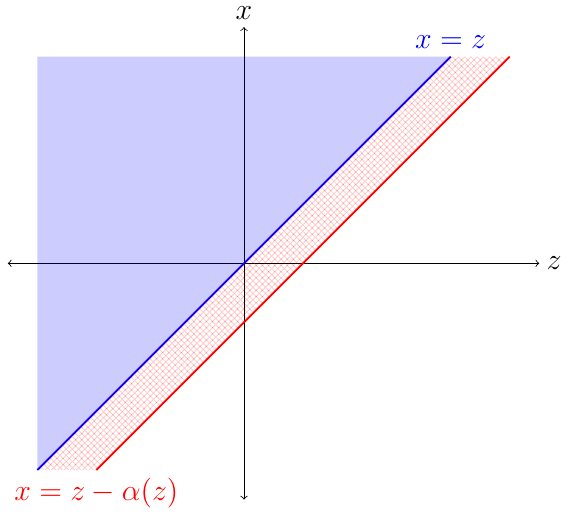}}
    \caption{We can calculate $\mathbb{E}[\severity \mid \eventininterval=0]$ by taking the expectation over the blue region and $\mathbb{E}[\severity \mid \eventininterval=0, \demographic=\othergroup]$ by taking the expectation over the blue plus crosshatched red regions, with each point having probability $p(z)f(x)$.}
    \label{fig:bias-proof-fig3}
    \end{center}
    \vskip -.2in
\end{figure}

\FloatBarrier

We thus see that at each value of $x$, the red crosshatched region adds strictly positive weight to higher values of $z$; thus, the expectation of $z$ over the red crosshatched region \textit{plus} blue region must be higher than the expectation of $z$ over the blue region itself. We therefore conclude that $\mathbb{E}[\severity \mid \eventininterval=0, A = a] > \mathbb{E}[\severity \mid \eventininterval=0]$. 

It is clear to see that this argument extends naturally to show that if a group tends to come in \textit{more frequently} than the rest of the population, their severity will be overestimated: if for some group $\tilde{a}$ there is some positive $\alpha(z)$ for all $z$ such that $P(\eventininterval=1\mid\severity=z, \demographic=\tilde{\othergroup}) = P(\eventininterval=1\mid\severity=z + \alpha(z))$, it will hold that $\mathbb{E}[\severity \mid \eventininterval,\demographic=\tilde{\othergroup}] < \mathbb{E}[\severity \mid \eventininterval]$. Hence any model that does not take into account demographic disparities in visit frequency will lead to biased estimates of severity.

\end{proof}

We finally show that our model's specific parameterization of visit rate satisfies the properties we assume and will, therefore, exhibit the bias we characterize in this Theorem. As described in \S \ref{sec:model}, we model a patient's visit rate using an inhomogeneous Poisson process characterized by visit rate $\lambda_t = \exp(\poissonintercept + \poissonseveritycoeff \cdot \severity + \poissongroupcoeff{\othergroup}$); for simplicity, we consider the two-group case and pin $\poissongroupcoeff{\tilde{\othergroup}}=0$ for one group; without loss of generality, we assume that $\poissongroupcoeff{\othergroup} < 0$ (group $\othergroup$ has a lower visit rate at the same severity). We show that when $\poissonseveritycoeff > 0$, 
this parameterization satisfies each of the more general assumptions in Theorem \ref{claim:proof-visit-disparity}---namely, $P(\eventininterval=1\mid\severity)$ is strictly monotone increasing in $\severity$; $\lim\limits_{\severity\rightarrow -\infty} P(\eventininterval=1\mid\severity) = 0$; $\lim\limits_{\severity\rightarrow \infty} P(\eventininterval=1\mid\severity) = 1$; and $P(\eventininterval=1\mid\severity=z, \demographic=\othergroup) = P(\eventininterval=1\mid\severity=z - \alpha(z))$---and the theorem's results thus apply to our specific parameterization.

$P(\eventininterval=1\mid\severity)$ is strictly monotone increasing in $\severity$ because the rate $\lambda_t = \exp(\poissonintercept + \poissonseveritycoeff \cdot \severity + \poissongroupcoeff{\othergroup})$ is monotone increasing for $\poissonseveritycoeff > 0$. Further, $\lim\limits_{\severity\rightarrow -\infty} \lambda(\severity) = 0$, meaning the expected number of visits in any discrete time period limits to $0$ and the probability of an event in any $[t_i, t_{i+1})$ limits to 0; similarly, $\lim\limits_{\severity\rightarrow \infty} \lambda(\severity) = \infty$, meaning that the probability of an event in $[t_i, t_{i+1})$ limits to 1.

Finally, we want to show that $P(\eventininterval=1\mid\severity=z, \demographic=\othergroup) = P(\eventininterval=1\mid\severity=z - \alpha(z))$ for all $z$, where $\alpha(z)$ is a positive function of $z$. For any $z$, let $F_a(z) = P(\eventininterval=1\mid\severity=z, \demographic=\othergroup)$; $0 \leq F_a(z) \leq 1$. Because $P(\eventininterval=1\mid\severity=z)$ is continuous in $z$ and increases from 0 to 1, by the intermediate value theorem there must exist some value $z - \alpha(z)$ for any $z$ such that $P(\eventininterval=1\mid\severity=z-\alpha(z)) = F_a(z) = P(\eventininterval=1\mid\severity=z, \demographic=\othergroup)$. Because $\poissongroupcoeff{\othergroup} < 0$, $P(\eventininterval=1\mid\severity=z, \demographic=\othergroup) <  P(\eventininterval=1\mid\severity=z)$. So $\alpha(z)$ must be positive, as desired.

\section{Simulations}
\label{app:simulations}

Our simulations show that, on synthetic data, our model accurately recovers true data-generating parameters, learns severity estimates that are well-calibrated with ground truth, and produces less biased estimates of severity than models that do not account for disparities. We describe the simulations in detail below, and all associated code can be found at \url{https://github.com/erica-chiang/progression-disparities}.

\paragraph{Data generation.} We generate synthetic datasets by drawing parameter values for each dataset from the prior distributions assumed by our model.  We generate simulated data for 1000 patients in each dataset, each of whom is assigned to one of two demographic groups (50\% chance of being from either group). Our model priors are as follows (where the normal distribution is represented as $\mathcal{N}(\mu, \sigma)$, and $\mathcal{TN}(\mu, \sigma, a)$ indicates a normal distribution with a lower bound of $a$). 

As described in the main text, we pin values $\avginitseverityplain = 0$, $\stdinitseverityplain=1$, and $\poissongroupcoeffplain=0$ for one group, for identifiability. Then for the non-pinned group:
\begin{align*}
    \avginitseverityplain&\sim\mathcal{N}(0, 4)\\
    \stdinitseverityplain&\sim\mathcal{TN}(1, 0.1, 0)\\
    \poissongroupcoeffplain&\sim\mathcal{N}(0, 2) \\
\end{align*}

The remaining group-independent priors are:
\begin{align*}
    \avgrateplain&\sim\mathcal{N}(1,4)\\
    \stdrateplain&\sim\mathcal{TN}(0.1, 0.4, 0) \\
    \featurescale_0&\sim\mathcal{TN}(1,1, 0.5) \text{, to enforce positive constraint}\\
    \featurescale_i&\sim\mathcal{N}(0, 2) \text{, for $i>0$}\\
    \featureintercept&\sim\mathcal{N}(0,1)\\ \psi&\sim\mathcal{TN}(5, 1, 0)\\
    \poissonintercept&\sim\mathcal{N}(1.5, 0.1)\\
    \poissonseveritycoeff&\sim\mathcal{TN}(0.5, 0.1, 0.1)\\ 
\end{align*}

\begin{figure}[ht]
    \begin{center}
    \centerline{\includegraphics[width=0.8\columnwidth]{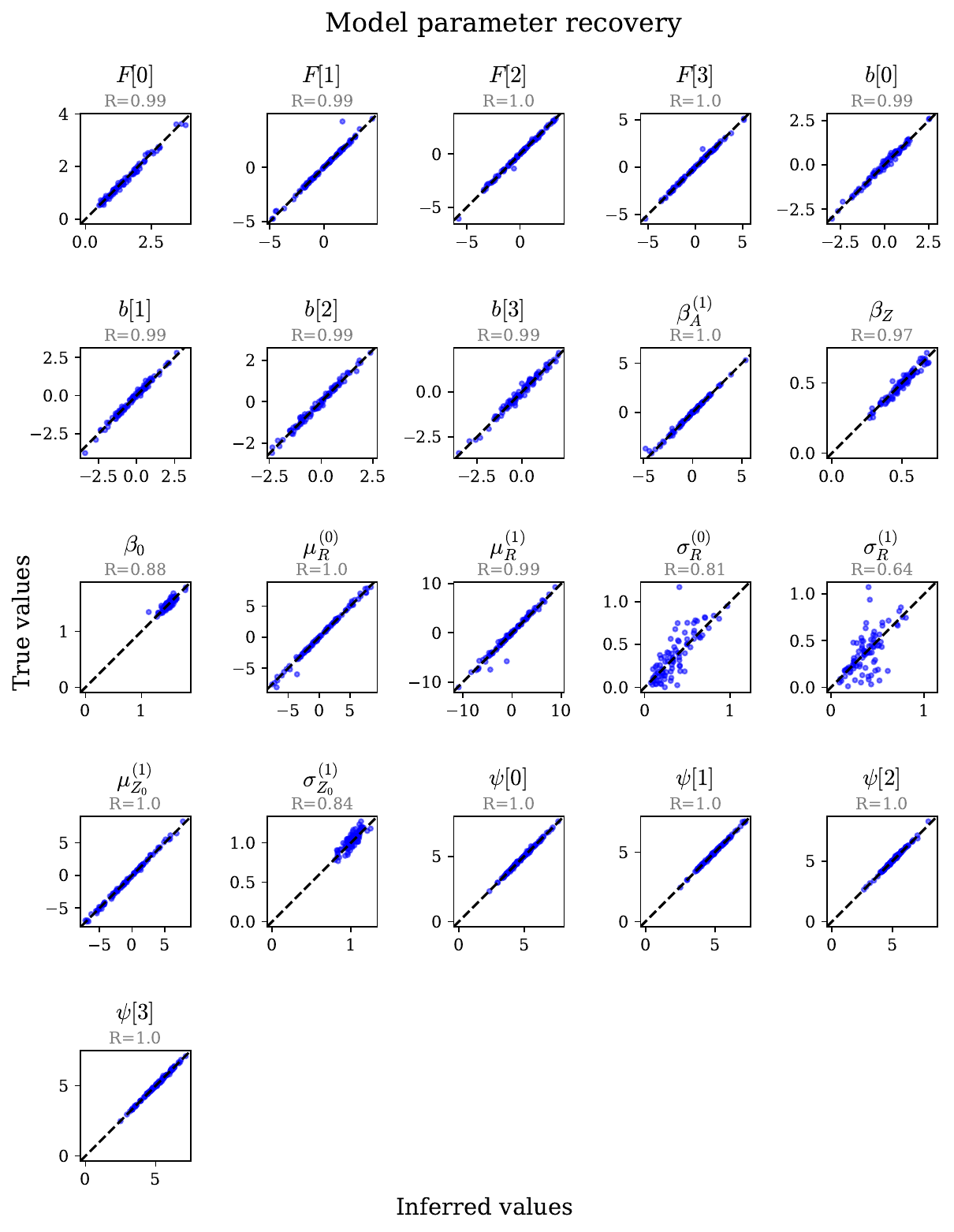}}
    \caption{\textbf{Parameter recovery from fitting our model to synthetic data.}}
    \label{fig:param_recovery}
    \end{center}
    \vskip -.2in
\end{figure}

\paragraph{Parameter recovery.} We fit our full model on $100$ synthetic datasets and compare the true data-generating values and recovered values of each parameter in our model. In Figure \ref{fig:param_recovery}, we visualize the recovery of each parameter by plotting true parameter values versus recovered posterior mean values, with one dot per run. 

\paragraph{Severity recovery.} We also compare the latent severity values of each patient at each timepoint to the recovered posterior mean values of severity for each patient. We examine the correlation between true and recovered values across both groups.

\section{NewYork-Presbyterian (NYP) Heart Failure Data Processing}
\label{app:real-data}
This study was conducted in accordance with Health Insurance Portability and Accountability Act (HIPAA) guidelines and with Institutional Review Board (IRB) approval.

\paragraph{Cohort filtering.} 
We analyze patients with \emph{heart failure with reduced ejection fraction} (HFrEF) whom we identify, following clinical guidance, by filtering the available NYP data for patients who have at least one LVEF measurement below 50\% and who have been recorded as receiving a diuretic prescription. To ensure we have relatively complete records for each patient, we then filter for patients who are likely to receive most of their cardiology care within the NYP system, by filtering for patients whose home zipcode is in the New York metropolitan area and who have at least two LVEF or BNP records at least 6 months apart within our data. Lastly, NYP switched electronic health record (EHR) systems, introducing inconsistencies in record-keeping across sites and years; to ensure our records are consistently recorded, we analyze data from Weill Cornell Medical Center, one of NYP’s two largest sites, between January 1, 2012 (the start of reliable record-keeping) to October 2, 2020 (NYP Cornell’s transition to a new EHR). This ensures records are consistently recorded in our data.

\paragraph{Feature processing.}
We convert pBNP to BNP with the conversion pBNP = 6.25 * BNP \citep{rorth_2020} and then log-transform BNP values to get one combined $\log_2$(BNP) feature \citep{hendricks_higher_2022}. We then normalize (z-score) all feature values so that each feature has mean $0$ and variance $1$. Because patient blood pressure and heart rate are much more likely to be measured at hospital visits unrelated to heart failure (while visiting another specialist in the NYP medical system), we limit patient observations to visits where a patient had one measurement of at least one of LVEF and BNP. 

We encode demographic categories by making $\demographic$ a one-hot encoding of race/ethnicity groups. Lastly, we describe the time scale of our model. As mentioned in \S\ref{sec:case_study}, we discretize time in 1-week bins; if a patient has multiple measurements of one feature within a timestep, we average all measurements within that timestep. Discretizing time in this way allows us to capture more long-term changes rather than acute changes in patient status. We normalize time so that the total time range in our model is 0 to 1. The longest patient trajectory in our data is 446 weeks (timesteps), so we normalize timestep values so that they range from 0 to 1; we therefore have fractional, discrete time values, each representing one week as $1/446$ units of time. 

\section{Model Evaluation}
\label{app:model-evaluation}

\paragraph{Fitting model on real data.} We fit our model on real data using weakly informative priors: $\avginitseverityplain\sim\mathcal{N}(0, 1)$, $\stdinitseverityplain\sim\mathcal{TN}(1,1, 0)$, and $\poissongroupcoeffplain \sim\mathcal{N}(0, 1)$ for the non-pinned groups; $\avgrateplain\sim\mathcal{N}(0,1)$ and $\stdrateplain\sim\mathcal{TN}(1.5, 1, 0)$ for all groups; $\featureintercept\sim\mathcal{N}(0,1)$; $\psi\sim\mathcal{TN}(1, 0.5, 0)$; $\poissonintercept \sim\mathcal{N}(2.5, 1)$; $\poissonseveritycoeff \sim\mathcal{N}(0, 1)$. For $\featurescale$, $\featureintercept$, and $\Psi$, we set model priors using Factor Analysis: at $t=0$, we have $\severity = \initialseverity \sim \mathcal{N}(0, 1)$ for group $\pinnedgroup$, meaning the mapping between severity and features 
$$X_0 = \featurescale \cdot Z_0 + \featureintercept + \featurenoise$$
$$\featurenoise \sim \mathcal{N}(0, \varfeaturenoise)$$
captures a factor analysis model with factor loading matrix $\featurescale$ and diagonal covariance matrix $\varfeaturenoise$. We run factor analysis using feature measurements from the \textit{first timestep} of all White patients (our $\pinnedgroup$ group) and use the estimates of $\featurescale$ from Factor Analysis as the mean of our priors on $\featurescale$. We define the variance of our priors on $\featurescale$ to be 1, and we pin the sign of $F_{\text{LVEF}}$ to be negative for identifiability. Since we have no inherent value scale for what $\featurescale$ values should be, Factor Analysis allows us to fit the model on more substantiated priors for feature scaling factors.

We then fit the model and get the parameter estimates from $1000$ samples. For any time $t$, we can calculate an estimate of $\severity$ and $\features$ for each sample, based on the sample's parameter estimates; we then take the average over all samples to get a patient's estimate of $\severity$ and $\features$. In order to get actual feature value estimates, we can linearly transform $\features$ to undo the normalization for each feature and recover an estimate of each feature value at $t$. We can then use our model's estimates of $\severity$ and predicted feature values to analyze and evaluate our model's behavior.

\paragraph{Comparison to baselines.} We filter out patients who do not have at least three visits (since several of the baselines we fit require this many visits per patient, as we describe below), leaving a total of 1834 patients: 1118 White, 347 Black, 216 Hispanic, and 153 Asian patients. 

To evaluate our model's ability to reconstruct feature values, we compare our model to PCA and FA. PCA and FA require consistent dimensionality of the input data, so we fit all models on the first three visits for each patient. We train two variants of both PCA and FA: the first attempts to reconstruct patient \emph{visits} from a single latent dimension (analogous to $Z$ in our model), taking as input the $\features$ vector at one visit (4 features total) and representing it with a single latent component. The second variant attempts to reconstruct \emph{patient trajectories} from two latent dimensions (analogous to $Z_0$ and $R$ in our model), taking as input a concatenated vector of features $\features$ from the first three visits (12 features total) and representing it with two latent components. We impute missing values as the overall mean of the data for both PCA and FA, since these methods cannot naturally handle missing data.

To evaluate our model's ability to predict future feature values, we compare our model to last time-step, logistic regression, and quadratic regression. Unlike PCA and FA, these methods do not require consistent dimensionality in the input data, so we fit the models to the first three years of observed data. Last-timestep predicts all future feature values to be equal to the most recent feature value observed in the training data for that patient; if there is no observed feature value, the baseline predicts the population mean. Linear regression regresses values on time for each patient and each feature to predict future feature values. For patients with fewer than 2 observations for a given feature value, we use the population mean for the preceding or subsequent timestep. Quadratic regression follows a similar approach. Because linear regression and quadratic regression can overfit the data and make unrealistic predictions, we clip their predicted feature values to a range determined by that observed within the training data. 

\paragraph{Ablated Model.} We compare our full model to an ablated version of the model that does not account for any of our three disparities. We do this by removing all group-specific parameters from the model, while leaving everything else the same: we learn one value of $\avgrateplain$ and $\stdrateplain$ and exclude $\poissongroupcoeffplain$ from the model. Since the distribution of $\initialseverity$ must be fixed for at least one group for identifiability (to fix the scale of $\severity$), the distribution is pinned for all groups. Factor Analysis for model priors on $\featurescale$ is also fit on all patients rather than only on white patients.

\section{Disparities Estimates}
\label{app:disparities-estimates}
We first describe our calculations for \S\ref{subsec:disparities} to estimate how much later Black and Asian patients start receiving care for heart failure compared to White patients. Our model learns the following: 
\begin{align*}
    \avginitseverity{\text{Black}} = \avginitseverity{\text{White}} + 0.22\\
    \avginitseverity{\text{Asian}} = \avginitseverity{\text{White}} + 0.27\\
\end{align*}
The learned average rate of progression across all patients is $0.62$. This means that Black patients come in $0.22/0.62 = 0.35$ units of time later in their disease progression than White patients, and Asian patients come in $0.27/0.62 = 0.44$ units of time later than White patients. Given that one unit of time is the longest patient trajectory, $8.5$ years, this leads us to $3.0$ and $3.8$ years for Black and Asian patients, respectively. 

Next we describe our calculations to estimate how much less frequently Black patients visit the hospital than White patients at the same disease severity. Our model learns that $$\poissongroupcoeff{\text{Black}} = \poissongroupcoeff{\text{White}} - 0.11$$

At the same disease severity $\severity$, Black patients will have a visit rate of 
\begin{align*}
    \lambda_t& = \exp(\poissonintercept + \poissonseveritycoeff\cdot \severity + (\poissongroupcoeff{\text{White}} - 0.11))\\
    &= \exp(\poissonintercept + \poissonseveritycoeff\cdot \severity + \poissongroupcoeff{\text{White}}) \cdot \exp(-0.11)\\
    &= 0.897\cdot \exp(\poissonintercept + \poissonseveritycoeff\cdot \severity + \poissongroupcoeff{\text{White}})\\
\end{align*}

So at the same disease severity, we estimate that Black patients have a visit rate that is 90\% that of a White patient's visit rate.

\newpage
\FloatBarrier
\section{Supplementary Figures and Tables}
\label{app:figures}
\renewcommand{\arraystretch}{1.2}
\begin{table*}\centering
\small
\begin{tabular}{@{}lccccc@{}}\toprule
& \textbf{Our model} & \textbf{$\text{FA}_{\text{visit}}$} & \textbf{$\text{PCA}_{\text{visit}}$} & \textbf{$\text{FA}_{\text{patient}}$} & \textbf{$\text{PCA}_{\text{patient}}$}\\ 
\midrule
MAPE: informative & 20\% & 28\% & 23\% & 25\% & 21\%  \\
MAPE: all & 16\% & 19\% & 17\% & 18\% &16\%\\
\bottomrule
\end{tabular}
\caption{\textbf{Our model compared to standard baselines for reconstruction performance.} We compare to factor analysis and principal component analysis fit at the patient visit level ($\text{FA}_{\text{visit}}$, $\text{PCA}_{\text{visit}}$) and at the trajectory level ($\text{FA}_{\text{patient}}$, $\text{PCA}_{\text{patient}}$). Models are fit on the first 3 visits from each patient and evaluated on same data using mean absolute percentage error (MAPE). We report aggregate performance for features that are more informative of heart failure progression (LVEF and BNP), along with performance for all features (LVEF, BNP, systolic blood pressure, heart rate).}
\label{table:reconstruction}

\vskip 0.8in

\begin{tabular}{@{}lccccc@{}}\toprule
& \textbf{Our model} & \textbf{Linear regression} & \textbf{Quadratic regression} & \textbf{Latest timestep} \\ 
\midrule
MAPE: informative & 28\% & 39\% & 59\%  & 22\%  \\
MAPE: all & 21\%  & 32\% & 49\% & 18\% \\
\bottomrule
\end{tabular}
\caption{\textbf{Our model compared to standard baselines for predictive performance.} We compare to linear regression, quadratic regression, and latest timestep prediction, each fit at the patient feature level. Models are fit on data from the first 3 years of each patient's disease trajectory and evaluated on visits after 3 years using mean absolute percentage error (MAPE). We report performance for features that are more informative of heart failure progression (LVEF and BNP), along with performance for all features (LVEF, BNP, systolic blood pressure, heart rate).}
\label{table:prediction}
\end{table*}

\begin{figure}[ht]
    \vskip -.2in
    \begin{center}
    \centerline{\includegraphics[width=0.7\columnwidth]{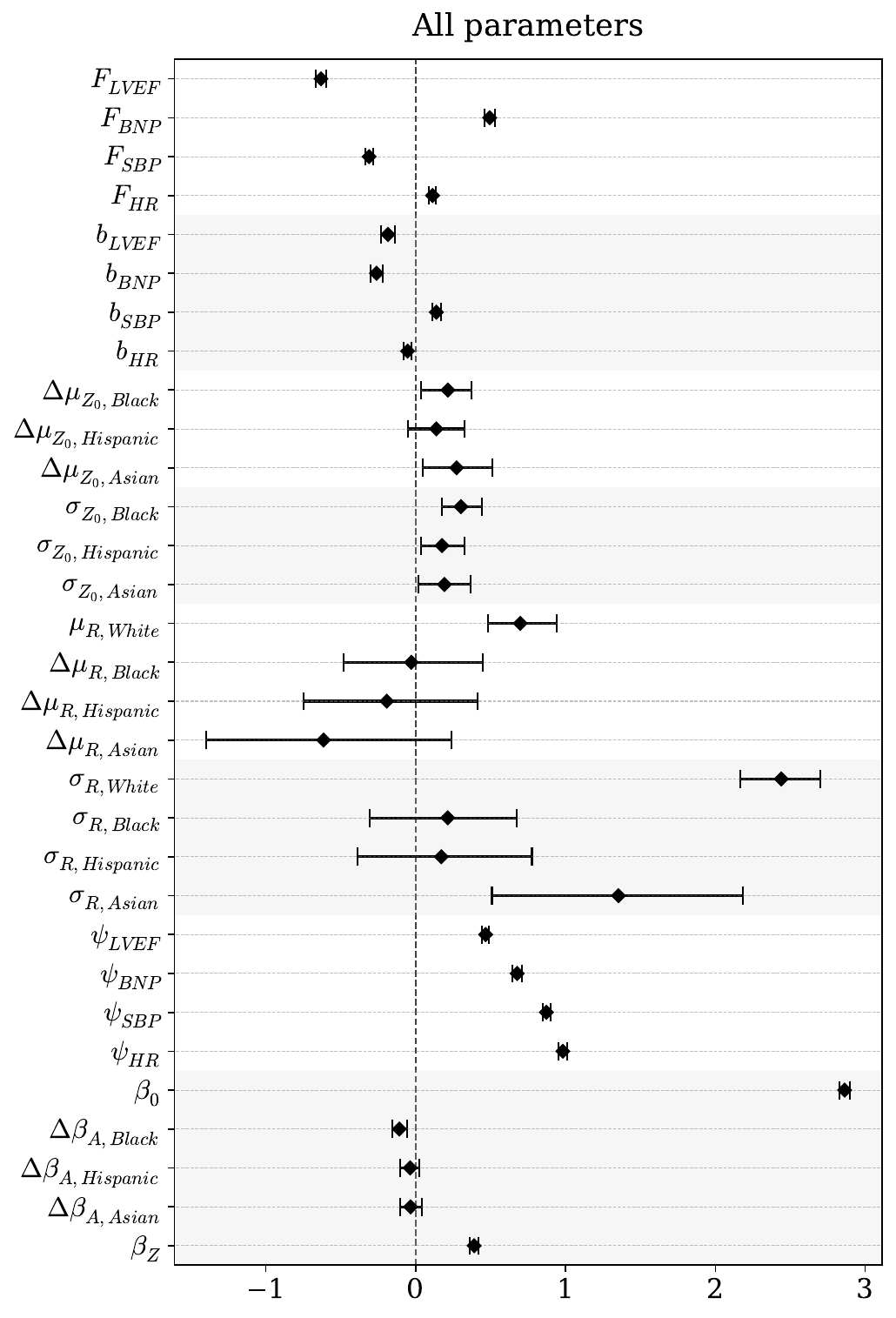}}
    \caption{\textbf{All parameters learned from fitting model on heart failure cohort.} Parameters of primary interest for interpreting our model (a subset of the parameters shown here) are highlighted in Figure \ref{fig:coeff_plot}.}
    \label{fig:all_coeff_plot}
    \end{center}
    \vskip -.2in
\end{figure}

\end{document}